\newif\ifarxiv \arxivtrue
\ifarxiv
\documentclass[acmtog, authorversion, nonacm]{acmart}
\else
\documentclass[acmtog,anonymous,review]{acmart}
\acmSubmissionID{318}  %
\fi

\usepackage{booktabs} %

\usepackage{algorithm}%
\usepackage{algpseudocode}%

\acmJournal{TOG}

\ifundef{\ifdraft}     {\newif\ifdraft \draftfalse}         {}
\drafttrue
\ifdraft
\newcommand{\gtc}[1]{\textcolor{cyan}{GT: #1}}

\newcommand{\gtdel}[1]{\textcolor{cyan}{\st{#1}}}

\newcommand{\abc}[1]{\textcolor{purple}{AB: #1}}
\newcommand{\ysc}[1]{\textcolor{orange}{YS: #1}}

\newcommand{\ysdel}[1]{\textcolor{orange}{\st{#1}}}

\newcommand{\rkc}[1]{\textcolor{olive}{RK: #1}}
\newcommand{\rkdel}[1]{\textcolor{olive}{\st{#1}}}

\else
\newcommand{\gtc}[1]{}

\newcommand{\gtdel}[1]{}
\newcommand{\abc}[1]{}
\newcommand{\ysc}[1]{}

\newcommand{\ysdel}[1]{}
\newcommand{\rkc}[1]{}

\newcommand{\rkdel}[1]{}
\fi

\usepackage{multicol}
\usepackage{multirow}
\usepackage{soul}
\usepackage{lipsum}

\newcommand\blfootnote[1]{%
  \begingroup
  \renewcommand\thefootnote{}\footnote{#1}%
  \addtocounter{footnote}{-1}%
  \endgroup
}

\begin{document}
\title{Human Motion Diffusion as a Generative Prior}

\author{Yonatan Shafir$^*$}
\author{Guy Tevet$^*$}
\author{Roy Kapon}
\author{and Amit H. Bermano} 
\affiliation{%
  \institution{Tel-Aviv University}
  \country{Israel}
}
\email{{Shafir2,guytevet}@mail.tau.ac.il}

\begin{teaserfigure}
\centering
\large
\vspace{10pt}
\includegraphics[width=\columnwidth]{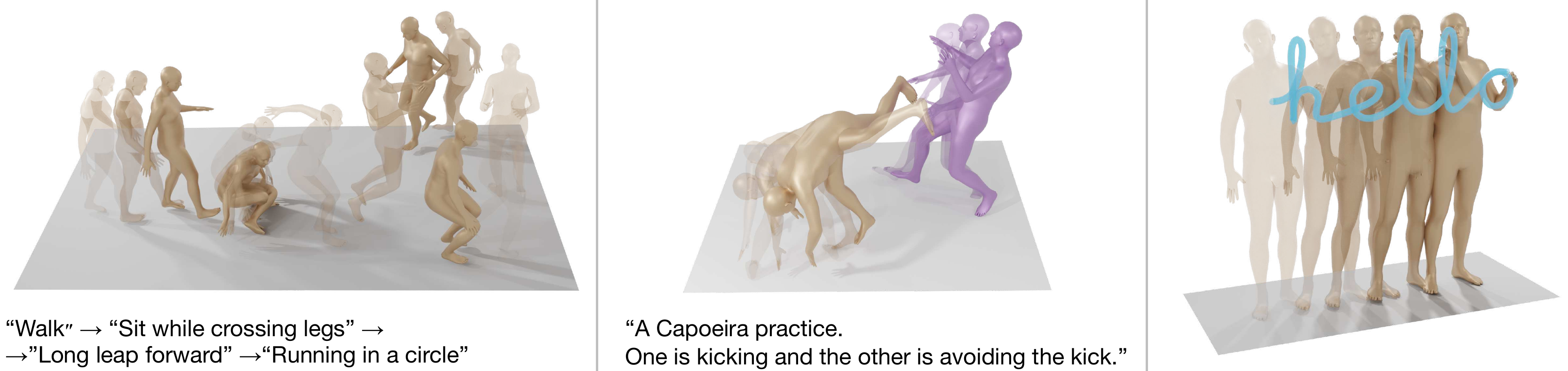}
\vspace{10pt}
\caption{
We suggest three novel motion composition methods, all based on the recent Motion Diffusion Model (MDM). 
\textbf{(Left) Sequential composition} generating an arbitrary long motion with text control over each time interval. 
\textbf{(Middle) Parallel composition} generating two-person motion from text. A different color represents a different person - both are generated simultaneously given the text prompt.
\textbf{(Right) Model composition} achieving accurate and flexible control by blending models with different control signals  - here writing ``hello" in mid-air.
}
\vspace{10pt}
\label{fig:teaser}
\end{teaserfigure}

\begin{abstract}

Recent work has demonstrated the significant potential of denoising diffusion models
for generating human motion, including text-to-motion capabilities.
However, these methods are restricted by the paucity of annotated motion data,
a focus on single-person motions, and a lack of detailed control.
In this paper, we introduce three forms of composition based on diffusion priors:
sequential, parallel, and model composition.
Using sequential composition, we tackle the challenge of long sequence
generation. We introduce DoubleTake, an inference-time method with which
we generate long animations consisting of sequences of prompted intervals
and their transitions, using a prior trained only for short clips.
Using parallel composition, we show promising steps toward two-person generation.
Beginning with two fixed priors as well as a few two-person training examples, we learn a slim
communication block, ComMDM, to coordinate interaction between the two resulting motions.
Lastly, using model composition, we first train individual priors
to complete motions that realize a prescribed motion for a given joint.
We then introduce DiffusionBlending, an interpolation mechanism to effectively blend several
such models to enable flexible and efficient fine-grained joint and trajectory-level control and editing.
We evaluate the composition methods using an off-the-shelf motion diffusion model,
and further compare the results to dedicated models trained for these specific tasks.
\url{https://priormdm.github.io/priorMDM-page/}
\footnote{Our code and trained models are available at \url{https://github.com/priorMDM/priorMDM}.}
\blfootnote{* The authors contributed equally}

\end{abstract}

\maketitle

\section{Introduction}
\label{sec:intro}

Human Motion Generation has recently experienced a tremendous leap forward. The recent elaborate language models~\cite{radford2021learning,devlin-etal-2019-bert} and diffusion generation approach~\citep{sohl2015deep,ho2020denoising} have quickly found their way into the field, yielding motion generation models that produce diverse and high-quality sequences from text or other forms of control~\cite{tevet2023human,tevet2022motionclip,petrovich22temos,guo2022generating}. 
In turn, these models have been already applied in the world of gaming, and hold the potential to open the field of character animation to novices and professionals alike.

However, the main problem the field of human motion generation has always struggled with and is still struggling with is data. Motion data is typically either acquired by elaborate motion capture settings~\cite{Joo_2015_ICCV} or crafted by artists~\cite{mixamo}. Both cases eventually lead to expensive and relatively small and homogeneous datasets~\cite{BABEL:CVPR:2021,guo2022generating}. For example, the datasets that current models are trained on, consist almost exclusively of short, single-person sequences. In the absence of data, tasks like multi-person interaction and long sequence generation are left behind, with poor generation quality. 

In this paper, we show that pretrained diffusion-based motion generation models can be leveraged as priors for composition, allowing out-of-domain motion generation and efficient control. 
Contrary to the high data consumption reputation of diffusion models, we show three methods that overcome the cost barrier using the aforementioned prior, enabling non-trivial tasks in few-shot or even zero-shot settings. %

In particular, we choose a pretrained Motion Diffusion Model (MDM)~\cite{tevet2023human} to serve as the prior. MDM achieves state-of-the-art results in the text-to-motion and action-to-motion tasks for short single-person sequences, and has already been demonstrated to generalize well to conditions from other domains~\cite{tseng2022edge}, and to corrections performed between the sampling iterations~\cite{yuan2022physdiff}.

Using this prior, we demonstrate three forms of composition: 
\begin{itemize}
    \item \textbf{Sequential composition}, where short sequences are concatenated to create a single long and coherent motion;
    \item \textbf{parallel composition}, where two single motions are coordinated to perform together;
    \item and \textbf{model composition}, where the motions generated by models with different control capabilities are blended together for composite control.
\end{itemize} 
Our DoubleTake method (Figure~\ref{fig:teaser}-Left), suggests a \emph{sequential composition} by carefully composing two generated motions in time, including the transition between them, and enables the efficient generation of long motion sequences in a zero-shot manner. Using it, we demonstrate $10$-minute long fluent motions that were generated using a model that was trained only on up to ~$10$ seconds long sequences~\cite{guo2022generating,BABEL:CVPR:2021}.
In addition, due to the composite nature of the generation, DoubleTake allows individual control for each motion interval, while maintaining consistent motion and transitions. This result is fairly surprising considering that such transitions were not explicitly annotated in the training data. 
DoubleTake consists of two phases for every diffusion iteration - in the first step, the individual motions, or intervals, are generated together in the same batch, each aware of the context of its neighboring intervals. Then, the second take refines the transitions between intervals to better match those generated in the previous phase.

For \emph{parallel composition}, we consider a few-shot setting, and enable textually driven two-person motion generation for the first time (Figure~\ref{fig:teaser}-Middle). Using our prior-based approach, we demonstrate promising two-person motion generation using only as few as a dozen training examples. The key idea is that in order to learn human interactions, we only need to enable prior models to communicate with each other throughout the diffusion process. Hence, we learn a slim communication block, ComMDM, that passes a communication signal between the two frozen priors through intermediate activation maps. 

Finally, we introduce a novel control mechanism via \emph{model composition}. We observe that the motion inpainting process suggested by \citet{tevet2023human} does not extend well to more elaborate yet important motion tasks such as trajectory and end-effector tracking. Hence, we first show that fine-tuning the prior for this task yields satisfying results while controlling even just a single end-effector. Then, we introduce the DiffusionBlending technique, which generalizes classifier-free guidance~\cite{ho2022classifier} to compose together different fine-tuned models and thus enables cross combinations of keypoints control on the generated motion. 
This enables surgical and flexible control for human motion that comprises a key capability for any animation system (Figure~\ref{fig:teaser}-Right).

We demonstrate, both quantitatively and qualitatively, that these inexpensive composition methods extend a more elaborately trained motion prior and outperform dedicated previous art in the respective tasks~\cite{wang2021multi,TEACH:3DV:2022}.

\section{Related Work}
\label{sec:rw}

\subsection{Motion Diffusion Models}

Very recently, MDM~\cite{tevet2023human}, MotionDiffuse~\cite{zhang2022motiondiffuse}, MoFusion~\cite{dabral2023mofusion}, and FLAME~\cite{kim2022flame} successfully implemented motion generation neural models using the Denoising Diffusion Probabilistic Models (DDPM)~\cite{ho2020denoising} setting, which was originally suggested for image generation. 
MDM enables both high-quality generation and generic conditioning that together comprise a good baseline for new motion generation tasks. 
EDGE~\cite{tseng2022edge} followed MDM by extending it for the music-to-motion task. 
SinMDM~\cite{raab2023single} adapted MDM to non-human motions using a single-sample learning scheme.
PhysDiff~\cite{yuan2022physdiff} added to MDM a pre-trained physical model based on reinforcement learning which enforces physical constraints during the sampling process.
These examples demonstrate the flexibility of MDM to novel tasks. 

In the images domain, \citet{Rombach_2022_CVPR} 
observed that training a model specifically for the inpainting task improves results. They input the inpainting mask as an additional control signal. \citet{meng2022sdedit}, \citet{lugmayr2022repaint}, and \citet{9711284} 
suggested various diffusion image editing methods based on partial denoising.

\subsection{Long-Sequence Motion Generation}

Motion Graphs~\cite{kovar2008motion} can synthesize long motions via traversing discrete poses given a data corpus. This approach is limited to existing data and will fail to generalize for elaborate textual conditions.
RNN-based motion generation tends to collapse into constant poses. \citet{martinez2017human} and \citet{zhou2018auto} overcome this issue  by feeding the model with its own generated frames during training for the task of prefix completion. Yet, those methods are still limited to the relatively short sequences of the available data. 
More recently, several works suggested breaking the data limitation by auto-regressively generating short sequences each one conditioned on a textual prompt and the suffix of the previous sequence. Transitions were either learned according to a smoothness prior~\cite{TEACH:3DV:2022, mao2022weakly} or from data~\cite{TEACH:3DV:2022,wang2022neural}, using the BABEL dataset~\cite{BABEL:CVPR:2021}, which explicitly annotates transitions between actions. EDGE~\cite{tseng2022edge} suggested the unfolding method to generate long sequences with SLERP interpolating between every two neighboring sequences. Contrarily, our DoubleTake suggests an unfolding method that leverages diffusion and blends the motion together at each denoising step. 

\subsection{Multi-Person Motion Generation}
Data scarcity is a major obstacle for multi-person motion generation, and the number of works is limited accordingly.
MuPoTS-3D dataset~\cite{mehta2018single} includes 20 real-world multi-person sequences; CMU-Mocap~\cite{CMU-Mocap} and 3DPW~\cite{von2018recovering} includes $55$ and $27$ two-person motion sequences respectively. 
\citet{yin2018sampling} suggested overcoming the data barrier by exploiting 2D information.
Recently, \citet{song2022actformer} contributed the synthetic multi-person GTA Combat dataset.
None of the datasets is textually (or otherwise) annotated, hence, the recent MRT~\cite{wang2021multi} and SoMoFormer~\cite{vendrow2022somoformer} models learned the unsupervised prefix completion task. Both learned motions under the DCT transform, which promotes smoothness and unrealistic motion, although improving L2 error measures.
In this work, we textually annotate 3DPW and learn text-guided two-person motion generation for the first time.

\subsection{Human Motion Priors}

VPoser~\cite{SMPL-X:2019} is a human pose auto-encoder, trained on the AMASS motion capture dataset~\cite{AMASS:ICCV:2019}. It is used as a prior for motion applications, such as motion denoising, fitting SMPL~\cite{loper2015smpl} model to joint location and as a pose code book for motion generation~\cite{hong2022avatarclip}. More recently, \citet{tiwari2022pose} showed that such prior can be learned as an implicit model.
MoDi~\cite{raab2022modi} is an unsupervised motion generator, adapted from StyleGAN~\cite{karras2019style}. Without further training, it enables latent space editing and motion interpolation. 
Contrary to those examples, MotionCLIP~\cite{tevet2022motionclip} uses priors from the image and text domains to learn motion. It aligns the motion manifold with CLIP~\citep{radford2021learning} latent space. This enables inheriting the knowledge learned by CLIP to generate motions out of the data limitations.

In the diffusion context, MDM adapts diffusion image inpainting~\citep{song2020score,saharia2022palette} for motion editing applications. In this work, we extend this principle by solving non-trivial motion tasks 
in few to zero-shot settings. 
More recently, MLD~\cite{chen2022mld} 
learned a latent diffusion model, similar to LDM~\cite{rombach2022high}
, which enables generating motion latent code instead of the motion itself, and lets a larger and pre-trained motion generator translate it into the physical space.

\section{Method}
\label{sec:method}

\begin{figure}[t!]
\centering
\includegraphics[width=\columnwidth]{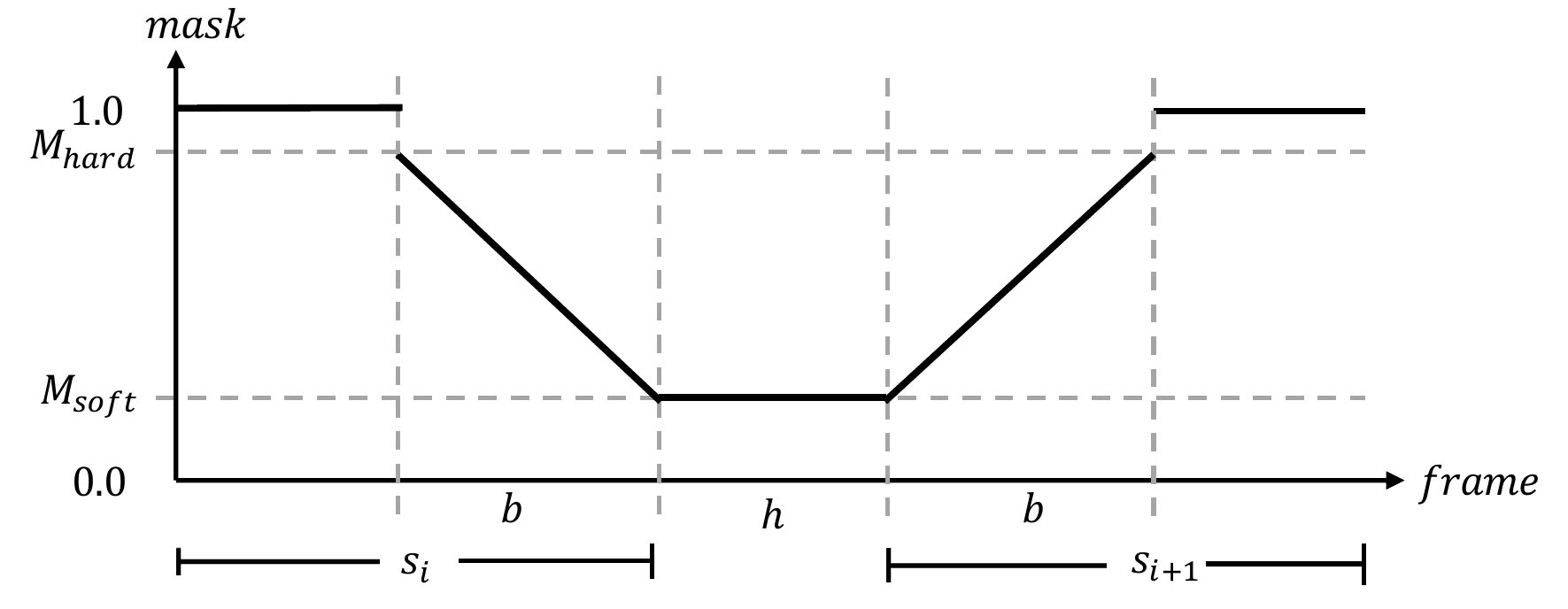}
\vspace{-15pt}
\caption{
\textbf{Soft blending overview.} We allow \textbf{b} frames long linear masking between $\mathbf{M_{hard}}$ to $\mathbf{M_{soft}}$ such that during the \textbf{Second take} at every denoising step part of the originally generated motion (suffix or prefix) going through refinement to fit the transition.}
\vspace{-10pt}
\label{fig:soft_b}
\end{figure}

\begin{figure}[t!]
\centering
\includegraphics[width=\columnwidth]{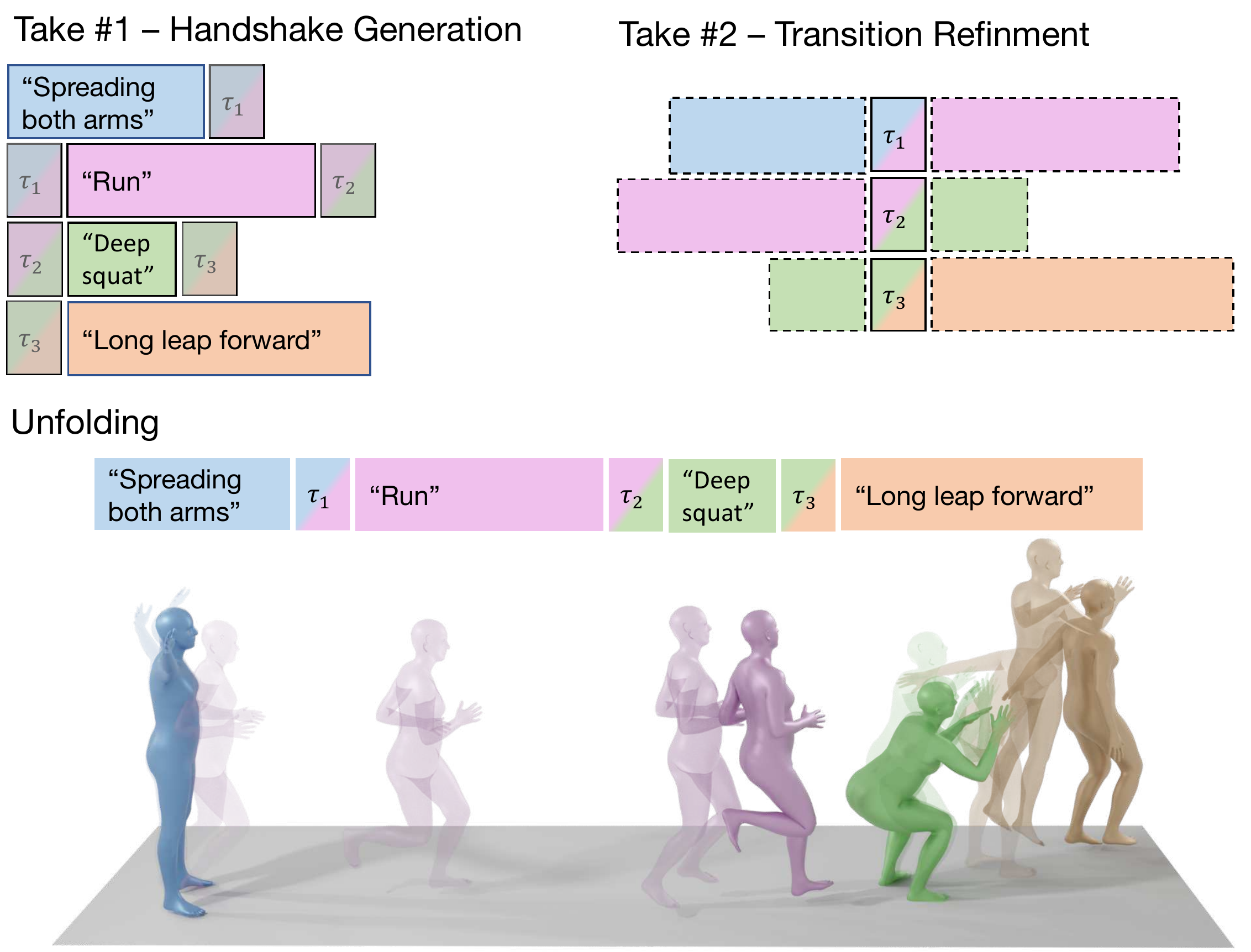}
\vspace{-15pt}
\caption{
\textbf{DoubleTake overview.} We generate arbitrarily-long sequences with text control per interval using a fixed motion diffusion prior. At the \textbf{first take}, we generate each interval as a single sample handshaking neighboring samples. At each denoising iteration, the handshakes are forced to be equal to eventually compose one long sequence. To refine the transition between intervals, the \textbf{second take} partially noise the handshakes and clean them conditioned on the neighboring intervals using a soft mask.
Solid frames mark generation or refinement; Dashed frames mark input motion to the take.
}
\vspace{-10pt}
\label{fig:double_take_overview}
\end{figure}

In this work, we use the recent Motion Diffusion Model (MDM)~\cite{tevet2023human}, pre-trained for the task of text-to-motion, to learn new generative tasks. We represent Human Motion as a sequence of poses $X=\{x^i\}_{i=1}^{N}$ where $x^i\in\mathbb{R}^{D}$ represent a single pose. 
Specifically, we use the SMPL~\cite{loper2015smpl} representation for experiments with the BABEL~\cite{BABEL:CVPR:2021} dataset, including joint rotations and global positions on top of a single human identity ($\beta = 0$). For all other experiments, we use the HumanML3D~\cite{guo2022generating} representation, composed of joint positions, rotations, velocities, and foot contact information. MDM is a denoising diffusion model based on the DDPM~\cite{ho2020denoising} framework. It assumes $T$ noising steps modeled by the stochastic process

\begin{equation}
{
q(X_{t} | X_{t-1}) = \mathcal{N}(\sqrt{\alpha_{t}}X_{t-1},(1-\alpha_{t})I),
}
\end{equation}

for a noising step $t \in T$, were $X_{T} \sim \mathcal{N}(0,I)$ is assumed. MDM models the denoising process: it predicts the clean motion $\hat{X}_{0}$ given a noised motion $X_t$, a noise step $t$ and a textual condition encoded to CLIP~\cite{radford2021learning} space and represented by $c$. 
The model is learned with the standard $\mathcal{L}_\text{simple} = E_{X_0 \sim q(X_0|c), t \sim [1,T]}[\| X_0 - MDM(X_t, t, c)\|_2^2]$ together with geometric losses that regulate the joint position, velocity and foot contact.
Sampling a novel motion from MDM is done in an iterative manner, according to \citet{ho2020denoising}. In every time step $t$  the clean sample $\hat{X}_{0}$ is predicted and noised back to $X_{t-1}$. This is repeated from $t=T$ until $X_0$ is achieved.

In this Section, we present \emph{sequential composition} with the DoubleTake method (\ref{sec:method_long}), which generalizes MDM to generate motions of arbitrary length without further training, through sequential composition. Then, we present \emph{parallel composition} by employing a slim communication layer, ComMDM (\ref{sec:method_multi}), trained with as few as $10$ interaction samples, for generating two-person motion. Lastly, we fine-tune MDM to control specific joints and present our \emph{model composition} method, DiffusionBlending (\ref{sec:method_edit}), that generalizes the classifier-free approach~\cite{ho2022classifier} to achieve fine-grained control over the body with any cross combination of joints to be controlled.

\subsection{Long Sequences Generation}
\label{sec:method_long}
Our goal is to generate arbitrarily long motions, such that each time interval of the motion is potentially controlled with a different text prompt and a different sequence length. We want the transitions between intervals to be realistic and to semantically match the neighboring intervals.
Since available datasets are limited in motion length and often do not explicitly include transitions, we suggest approaching this task in a zero-shot manner, using a fixed generative prior that was trained with such short sequences.
We present  DoubleTake (Figure~\ref{fig:double_take_overview}), a two-stage inference-time process that suggests a parallel solution and generates the long motion in a single batch. 
Typically, approaches that were designed specifically for this task ~\cite{TEACH:3DV:2022, mao2022weakly,wang2022neural} generate each such interval conditioned on the fixed suffix of the previous interval.
In contrast, 
DoubleTake generates a prompted interval while observing both the previous and next intervals, which are generated simultaneously.
In the first take, we generate each interval as a different sample in the denoised batch, such that each one is conditioned on its own text prompt and maintains a \emph{handshake} with its neighboring intervals through the denoising process.
Handshake, $\tau$, is defined as a short (about a second long) prefix or suffix  of the motion, such that the prefix of the current motion is forced to be equal to the suffix of the previous motion. 
Each interval maintains two such handshakes as demonstrated in Figure~\ref{fig:double_take_overview}. The handshake is maintained by simply overriding $\tau$ with the frame-wise average of the relevant suffix and prefix at each denoising step. 
This allows our model to generate long sequences that  depend on the past and future motions while being aware of the whole sequence during the generation of each interval. 
The handshake length $h = |\tau|$ can be arbitrarily defined by the user, also on a per-transition level. However, in practice, we find that the choice of one-second-long handshakes is robust throughout our experiments. 
Formally, handshakes are forced to be equal at the end of each denoising iteration as follows:
\begin{equation}
{
    \tau_i = (1-\vec{\alpha}) \odot S_{i-1}[-h:] + \vec{\alpha} \odot  S_i[:h]
}
\end{equation}
where $S_i$ indicates the $i^{th}$ sequence $\alpha_{j}={j}/{h}, \forall{j: j\in [0:h)}$
and $\odot$ indicates a element-wise multiplication. 

Looking at the generated handshaked motion however, we observe visually displeasing results, as artifacts and inconsistencies occur in the transitions between semantically different motions (i.e. ``Run" and then ``Crawl"). Consequently we suggest adding the \emph{second take}, applied on the output of the first take. In the second take, we reshape our batch as shown in Figure~\ref{fig:double_take_overview}, such that in each sample we get the transition sandwich ($S_i$, $\tau _{i}$, $S_{i+1}$).
Now, we partially noise the sandwich $T'$ noising steps and denoise it back to $t=0$ under our suggested \emph{soft-masking} feature to refine transitions:
In a regular inpainting mask, the content is either taken completely from the input, or is completely generated. We suggest a soft inpainting scheme, where each frame is assigned a soft mask value between $0$ and $1$ that dictates the amount of refinement the second take performs on top of the first take's result.
To this end, we define the masks $M_{soft}$, $M_{hard}$ for the interval $S$ and hanshake $\tau$ respectively, with a short, $b$ frames long,  linear transition between the mask values as demonstrated in Figure~\ref{fig:soft_b}. %

Finally, we construct the long sequence by \emph{unfolding it}, i.e. by reshaping each sequence and transition back to its linear place as 
demonstrated in Figure~\ref{fig:double_take_overview} bottom.

\begin{figure}[t!]
\centering
\includegraphics[width=\columnwidth]{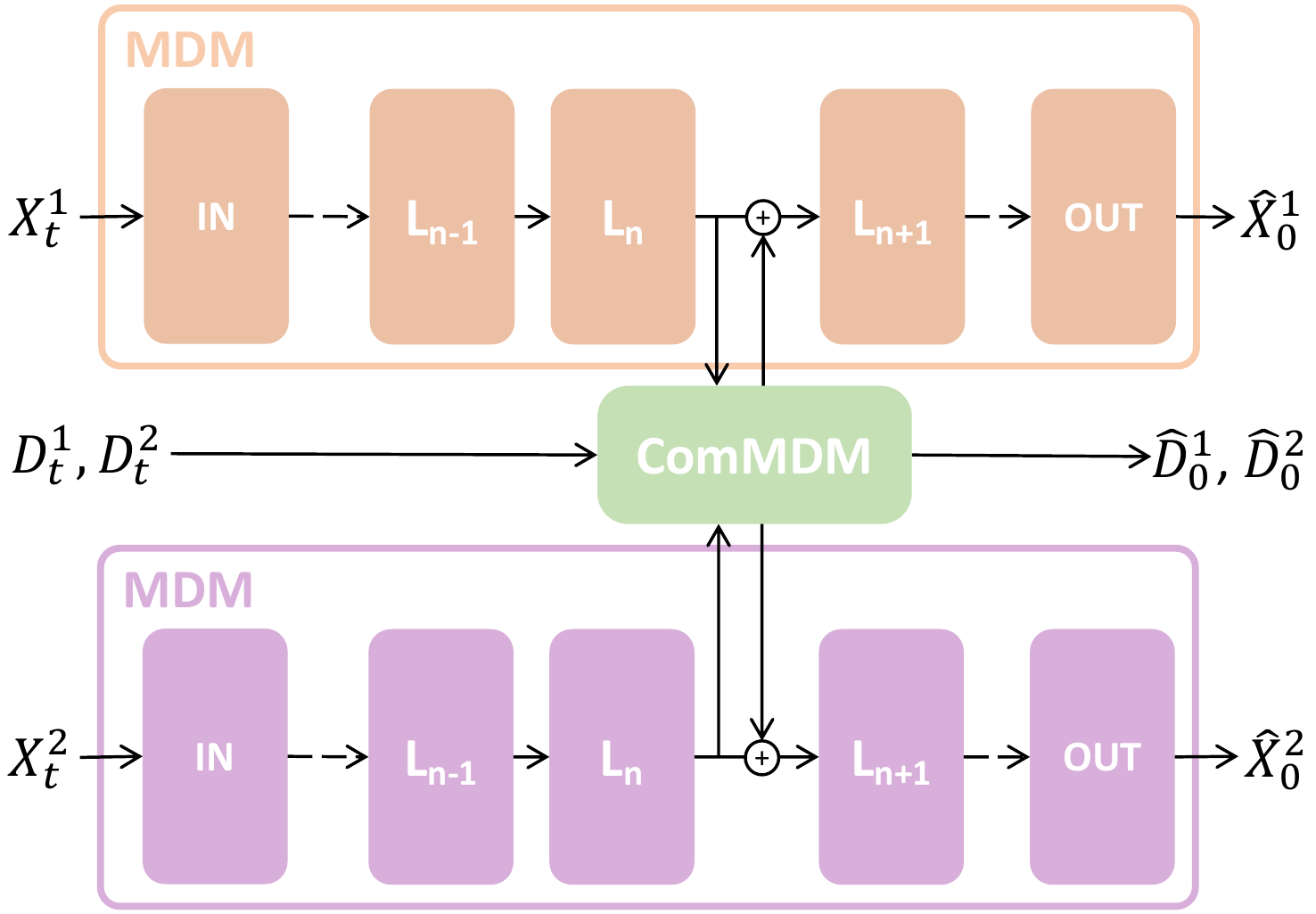}
\vspace{-15pt}
\caption{
\textbf{ComMDM overview.} Using two fixed MDM models, we train a slim communication block (ComMDM) for two-person motion generation. ComMDM gets as input the activations of transformer layer $L_n$ from both actors and outputs a correction term which is added to the same activations. Optionally, ComMDM also  predicts the initial poses $D^i$ of the two persons. IN and OUT stand for the linear input and output layers of the transformer.
}
\vspace{-10pt}
\label{fig:com_overview}
\end{figure}

\subsection{Two-Person Generation}
\label{sec:method_multi}

Our goal is to simultaneously generate motion of two people interacting with each other. The limited data availability dictates a few-shot learning solution. Our key insight is that by dedicating a fixed generator for each person in the scene the motion remains in the human motion distribution, and we only need to learn to coordinate between the two. 
Hence, we introduce ComMDM (Figure~\ref{fig:com_overview}), a single-layer transformer model that is trained to coordinate between two instances of a fixed MDM (one for each person). ComMDM is placed after transformer layer $n$, gets as inputs the output activations of this layer from both models $(O_{t}^{1,(n)}, O_{t}^{2,(n)})$ 
and outputs 
a correction term for each of the two models $\Delta O_{t}^{i,(n)}$.
To further reduce the number of learned parameters, we exploit symmetry considerations and output only one correction term, then the output to be corrected is entered first, such that the corrected output is
$\tilde{O}_{t}^{i,(n)} = O_{t}^{i,(n)} + ComMDM(O_{t}^{i,(n)}, O_{t}^{3-i,(n)})$. 
We note that in some datasets, such as HumanML3D, all motions are processed to start with the root at the origin and facing the same direction. 
Hence, naively using ComMDM on a model that was trained with such data will result in two people both being placed at the origin at the beginning of the motion. To mitigate that, ComMDM additionally learns $D$, the initial pose of each person at the first frame as a part of the diffusion process. Hence, the full implementation of ComMDM is $\Delta O_{t}^{i,(n)}, \hat{D}^i_0 = ComMDM(O_{t}^{i,(n)}, O_{t}^{3-i,(n)}, D^i_t, t)$.  

We freeze the weights of the MDM instances and train only ComMDM with the $L_\textrm{simple}$ loss. We learn two motion tasks; For prefix completion, we use a fine-tuned version of MDM for prefix completion (See~\ref{sec:method_edit}) and completely mask the textual condition. For the text-to-motion task, we use a regular instance of MDM and mask the textual condition with a probability of $10\%$ to support classifier-free guidance.

\subsection{Fine-Tuned Motion Control}
\label{sec:method_edit}
Our goal is to generate full-body motion controlled by a user-defined set of input features. These features can be root trajectory, a single joint, or any combination of them. We require a self-coherent generation that semantically adheres to the control signal.
For instance, when specifying the root trajectory of a person to move backward, we expect the generated motion to have the legs adjusted to walking backward.
As we show in \autoref{sec:Fine-Tuned Motion Control}, the motion in-painting method suggested by \citet{tevet2023human} fails to meet this requirement. 

\textbf{Single Control Fine-Tuning.} Consequently, inspired by \citet{Rombach_2022_CVPR}, we introduce a fine-tuning process to yield a model that adheres to the control features. In essence, our method works by masking out the noise applied to the ground-truth features we wish to control, during the forward pass of the diffusion process. This means that during training, the ground-truth control features propagate to the input of the model, and thus, the model learns to rely on these features when trying to reconstruct the rest of the features. Algorithm~\ref{alg:fine-tune} describes the fine-tuning process for trajectory control task.
For sampling, we follow the 
core idea of
the finetuning process: 
After we get the model's prediction of $x_0$, we inject the editing features into it. Then, in the forward process from the predicted $x_0$ to $x_{t-1}$, we mask out the noise in the control features 
to allow them to cleanly propagate into the model. 
Algorithm~\ref{alg:sampling} defines this sampling process for trajectory control task.
The fine-tuning stage requires less than $20K$ steps to generate visually pleasing results. 
It allows us to easily acquire a dedicated model for a given control task.

\begin{algorithm}[t]
\caption{Fine-tuning method}
\begin{algorithmic}
\Repeat
\State $x_0\sim q\left(x_0\right)$
\State $t\sim $Uniform$\left(\{ 1,\dots,T\}\right)$
\State $\epsilon\sim \mathcal{N}\left(0,I\right)$
\State $\epsilon\left[trajectory\right]=0$ \Comment{\textbf{Our addition}}
\State Take gradient descent step on:
\State \quad $\nabla_\theta\lVert x_0-\epsilon_\theta\left(\sqrt{\bar{\alpha_t}}x_0+\sqrt{1-\bar{\alpha_t}}\epsilon,t\right)\rVert$\;
 \Until{Converged}
\end{algorithmic}
\label{alg:fine-tune}
\end{algorithm}

\begin{algorithm}[t]
\caption{Sampling method}
\begin{algorithmic}
\State $x_0^{\left(T\right)}=0$
\For {$t=T,\dots,0$}
\State $x_0^{\left(t\right)}\left[trajectory\right] = $given trajectory \Comment{\textbf{Original in-painting}}
\State $\epsilon\sim \mathcal{N}(0,I)$
\State $\epsilon\left[trajectory\right]=0$ \Comment{\textbf{Our addition}}
\State $x_0^{\left(t-1\right)}=\epsilon_\theta\left(\sqrt{\bar{\alpha_t}}x_0+\sqrt{1-\bar{\alpha_t}}\epsilon,t\right)$
\EndFor
\end{algorithmic}
\label{alg:sampling}
\end{algorithm}

\textbf{DiffusionBlending.} 
A fine-tuned model for every possible control task is sub-optimal. 
Hence, we suggest DiffusionBlending, a \emph{model composition} method for using multiple models for composite control tasks. For instance, if we wish to dictate both the trajectory of the character and its left hand, we can blend the model that was trained solely for trajectory control and the model that was trained only for the left hand.

To control cross combinations of the joints (i.e. both the root and the end effector as in Figure~\ref{fig:teaser}), we extend the core idea of the classifier-free approach~\cite{ho2022classifier} and present DiffusionBlending. 
The classifier-free  approach suggests interpolating or extrapolating between the conditioned model $G$ and the unconditioned model $G^\emptyset$. %
We argue that this idea can be generalized to any two "aligned" (see definition in ~\cite{wu2021stylealign}) diffusion models $G^a$ and $G^b$ that 
are conditioned on $c_a$ and $c_b$ respectively. Then sampling with two conditions simultaneously is implemented as

\begin{equation}
{
G_s^{a,b}(X_t, t, c_a, c_b) = G^a(X_t, t, c_a) + s \cdot (G^b(X_t, t, c_b) - G^a(X_t, t, c_a)),
}
\end{equation}

with the scale parameter $s$ trading-off the significance of the two control signals.

\section{Experiments}
\label{sec:experiments}

\begin{table*}[ht]
\centering
\resizebox{0.9\textwidth}{!}{

\begin{tabular}{lcccc||cc||cc}
  \toprule
& \multicolumn{4}{c}{Motion} & \multicolumn{2}{c}{Transition (70 frames)} & \multicolumn{2}{c}{Transition (30 frames)} \\
\hline
& R-precision $\uparrow$ & FID$\downarrow$ & Diversity$\rightarrow$ & MultiModal-Dist$\downarrow$ &  FID$\downarrow$ & Diversity$\rightarrow$ &  FID$\downarrow$ & Diversity$\rightarrow$\\

\hline

Ground Truth  & ${0.62}$ & $0.4 \cdot 10^{-3}$ & ${8.51}$ & ${3.57}$ &  $0.8 \cdot 10^{-3}$ & ${8.23}$ & $0.9 \cdot 10^{-3}$ & ${8.33}$\\

\hline

\hline

TEACH~\shortcite{TEACH:3DV:2022} & $\underline{0.46}$ & $1.12$ & $\mathbf{8.28}$ & $7.14$ &  $3.86$ & $\mathbf{7.62}$& $7.93$ & $6.53$\\

\hline

Double Take (ours) & $ 0.43$ & $1.04$ & $8.14$& $7.39$ &  $\mathbf{1.88}$ & $7.00$ & $\mathbf{3.45}$ & $\mathbf{7.19}$\\

+ Trans. Emb 
& $\mathbf{0.48}$ & $\mathbf{0.79}$ & $\underline{8.16}$ & $\mathbf{6.97}$ & $3.43$ & $6.78$ & $7.23$ & $6.41$\\

+ Trans. Emb + geo losses 
& $0.45$ & $\underline{0.91}$ & $\underline{8.16}$& $\underline{7.09}$ &  $\underline{2.39}$ & $\underline{7.18}$ & $\underline{6.05}$ & $\underline{6.57}$\\

\bottomrule
\end{tabular} 
}
\caption{\textbf{Quantitative results on the BABEL~\shortcite{BABEL:CVPR:2021} test set.} All methods use the real motion length from the ground truth. `$\rightarrow$' means results are better if the metric is closer to the real distribution. We run all the evaluations 10 times. Transition metrics were tested on two different lengths and it contains context from the suffix of the previous frame and the prefix of the next frame. tested on two different margin lengths since TEACH~\shortcite{TEACH:3DV:2022} define transition of 8 frames. \textbf{Bold} indicates best result, $\underline{underline}$ indicates second best result. R-precision reported is top-3.}
\vspace{-20pt}
\label{tab:teachMDM}
\end{table*}

\begin{figure}[t!]
\centering
\includegraphics[width=\columnwidth]{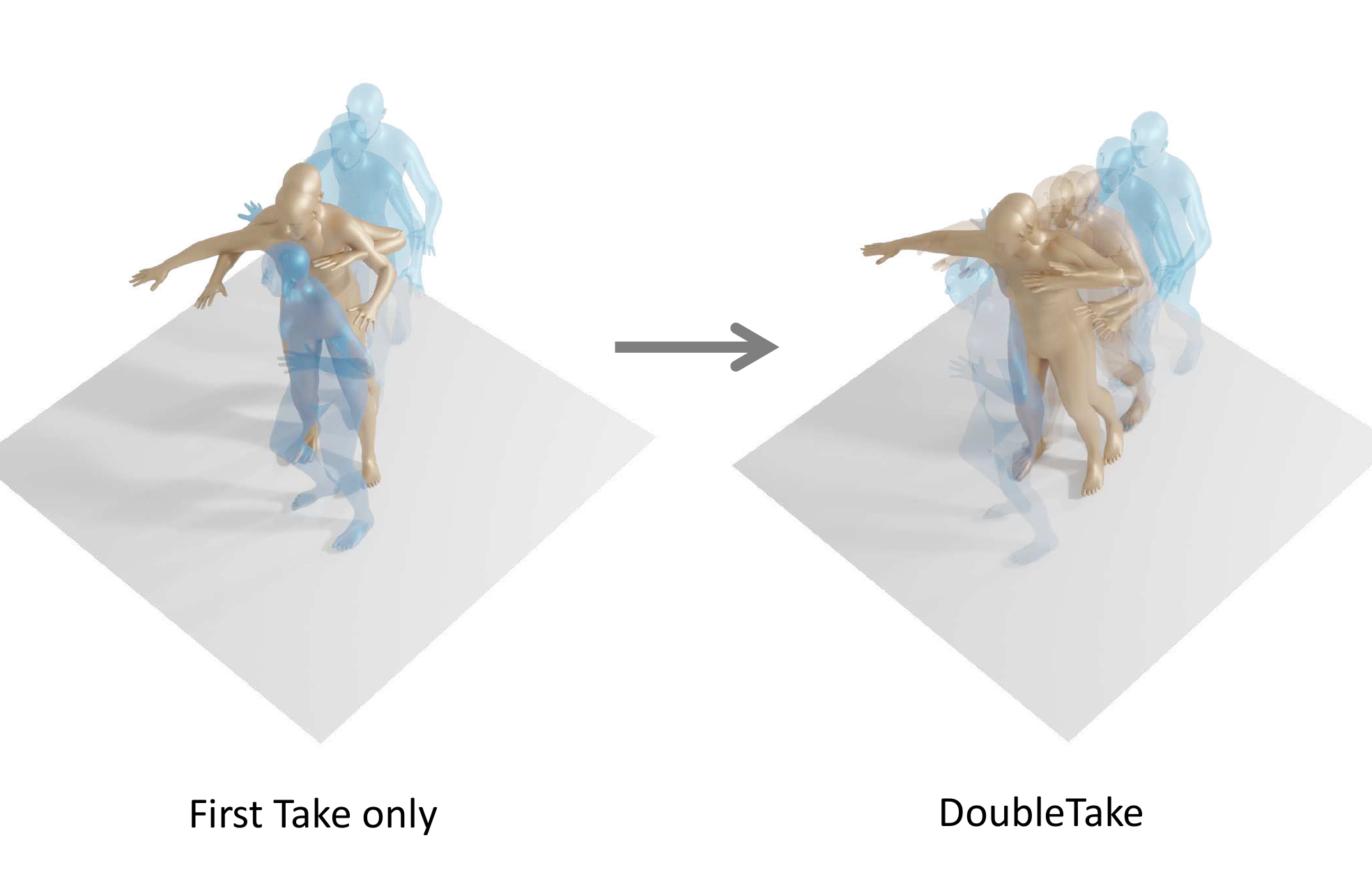}
\caption{
\textbf{DoubleTake transition refinement.} The second take refines the transitions generated in the first take to be more smooth and more realistic. Orange are subsequent transition frames and Blue are context intervals.
}
\label{fig:double_take_fail}
\end{figure}

\begin{figure}[t!]
\centering
\includegraphics[width=\columnwidth]{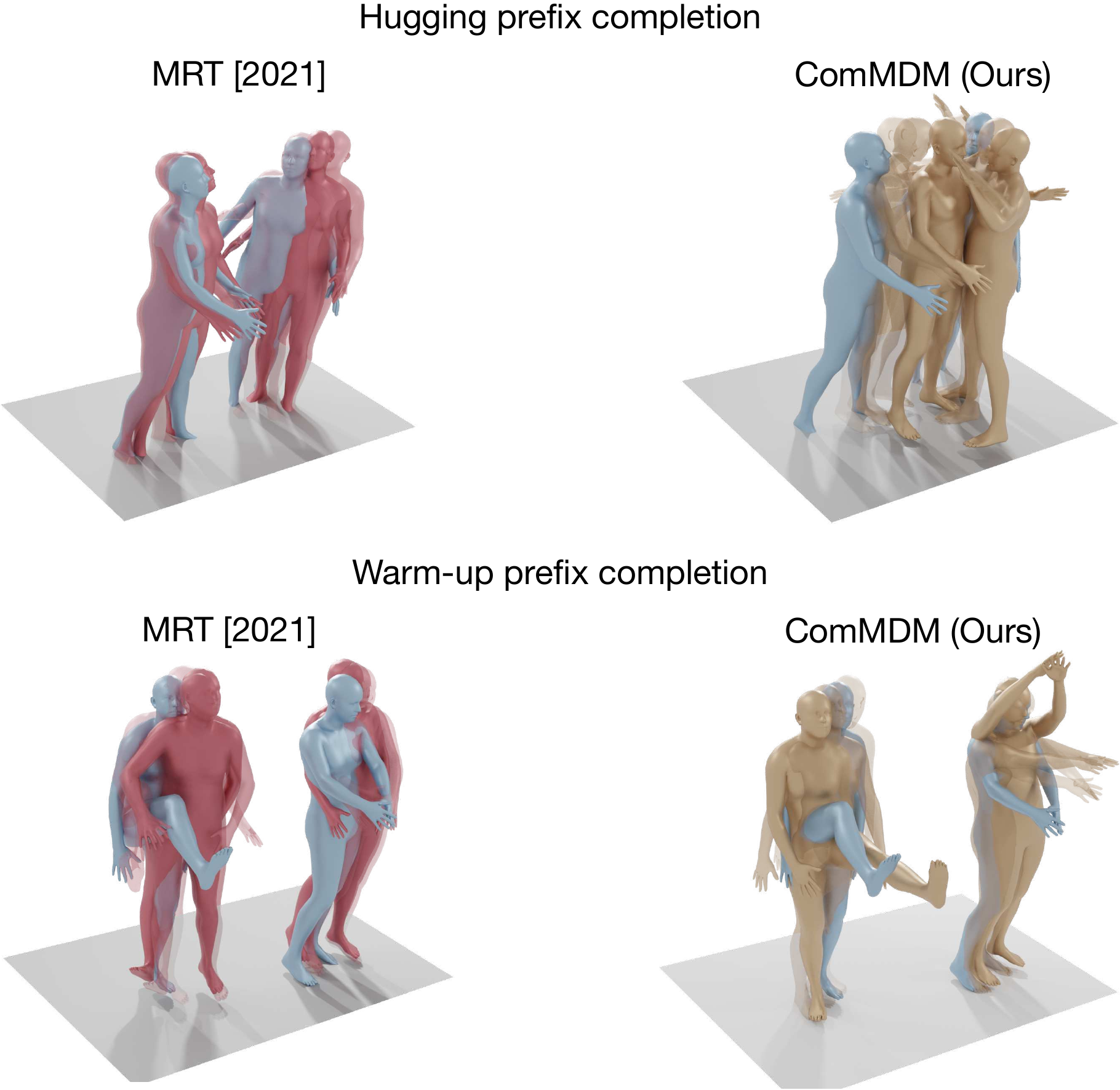}
\caption{
\textbf{Two-Person Prefix Completion.} MRT~\cite{wang2021multi} tends to fixate on the prefix pose whereas our ComMDM provides lively and semantically correct completions. Blue figures are the input prefix frames, provided to both models. The red and orange figures are MRT and our completions correspondingly.
}
\label{fig:two_person_comp}
\end{figure}

\subsection{Long Sequences Generation}

For long sequence generation with our DoubleTake method, we use a fixed MDM~\cite{tevet2023human} trained on the HumanML3D~\cite{guo2022generating} dataset, originally trained with up to $10$ seconds long motions.
To compare with TEACH~\cite{TEACH:3DV:2022}, which was dedicatedly trained for this task, we train MDM for $1.25M$ steps on BABEL~\cite{BABEL:CVPR:2021}, the same dataset TEACH was trained on with the same hyperparameters suggested by \citet{tevet2023human} on a single NVIDIA GeForce RTX 2080 Ti GPU. For both datasets, we applied DoubleTake with a one-second-long transition length,  $T'=700$,  $M_{hard}=0.85$, $M_{soft}=0.1$ and $b=10$.

In both cases, we evaluate the generation using the evaluators and metrics suggested by \citet{guo2022generating}. In short, they learn text and motion encoders for the HumanML3D dataset as evaluators that map motion and text to the same latent space, then apply a set of metrics on the generated motions as they are represented in this latent space. 
\emph{R-precision} measures the proximity of the motion to the text it was conditioned on, \emph{FID} measures the distance of the generated motion distribution to the ground truth distribution in latent space, \emph{Diversity} measures the variance of generated motion in latent space, and \emph{MultiModel distance} is the average $L2$ distance between the pairs of text and conditioned motion in latent space. For full details, we refer the reader to the original paper.
Note that for the BABEL dataset, we trained the same evaluators following the setting defined by \citet{guo2022generating}. 
To provide a proper analysis, we generate a 32-intervals long sequence, then apply HumanML3D metrics on the intervals themselves, and once again for the transition. Note that the text-related metrics are not relevant for transitions.

Since the BABEL dataset annotates transitions as well, we suggest using our \emph {Transition Embedding}: we choose to embed each frame with transition embedding signal, allowing the model better understand if the following frame belongs to transition or part of the motion. We then add this embedding to the frame's features. 
Additionally, we choose to train our model over the BABEL dataset with geometric losses as  proposed in MDM.
We note that whereas we do not apply any post-process to the motion, TEACH aligns the start of each interval to the end of the previous one and adds extra interpolation frames between the two. We observe that without this post-process TEACH produces poor transition, yet evaluated it with all the above to maintain fair conditions.

\begin{figure}[t!]
\centering
\includegraphics[width=\columnwidth]{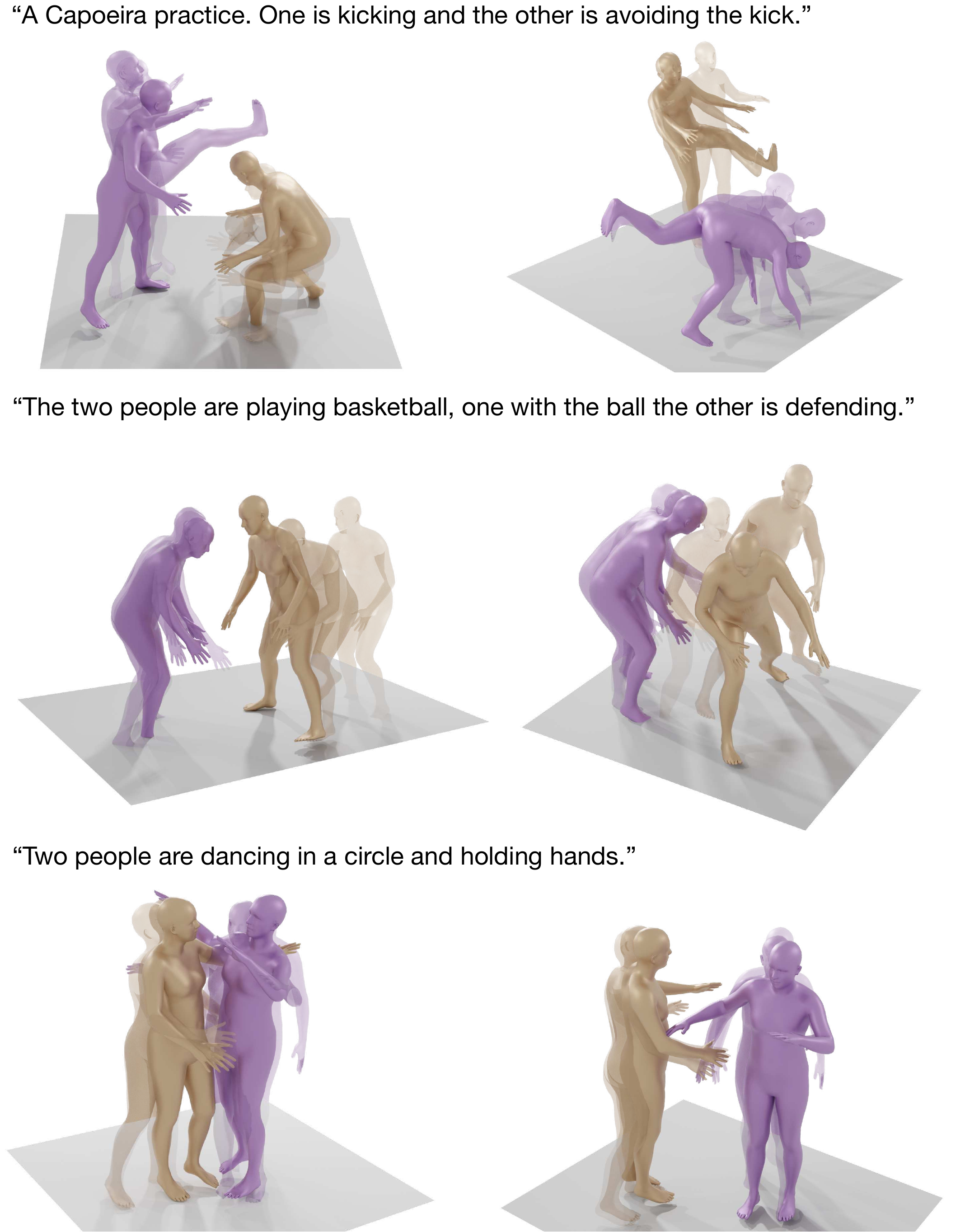}
\caption{
\textbf{Two-Person Text-to-Motion.} We use ComMDM to generate two-person interactions given an unseen text prompt describing it. Different color defines different character, both are generated simultaneously.
}
\label{fig:two_person_text}
\end{figure}

\begin{figure}[t!]
\centering
\includegraphics[width=\columnwidth]{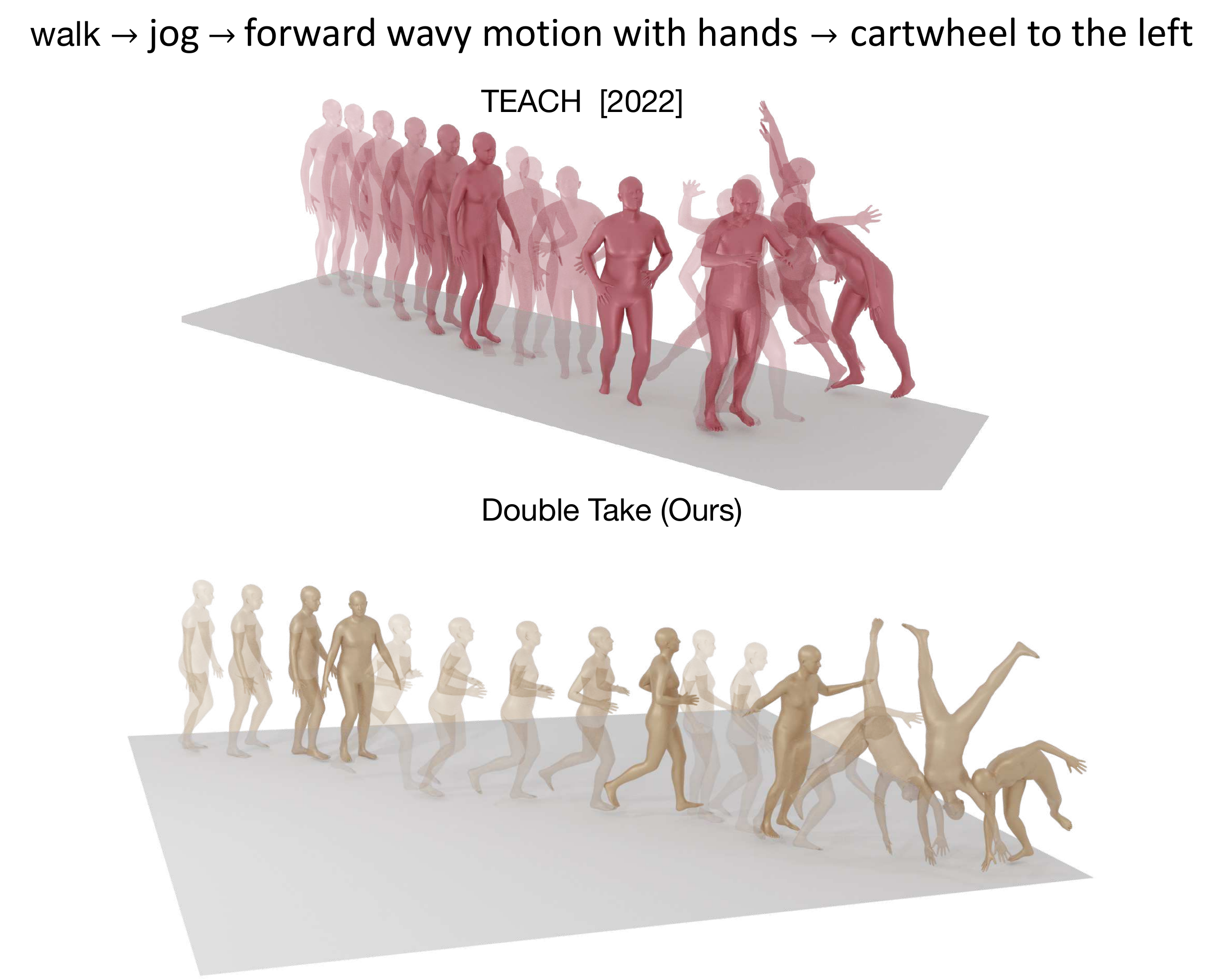}
\caption{
\textbf{DoubleTake compared to TEACH~\cite{TEACH:3DV:2022}.} We show that while our DoubleTake provides coherent motion with realistic transitions, TEACH generation suffers from sliding.
}
\label{fig:teach_compare}
\end{figure}

Table~\ref{tab:teachMDM} presents quantitative results over the BABEL dataset, compared to TEACH.
We evaluated the transitions with two variations - the first with fair margins from the intervals (70 frames) and the other with minimal possible margins for both DoubleTake and TEACH (30 frames which are 1 second). 
DoubleTake outperforms TEACH in terms of FID with all our methods. When considering transition evaluations the gaps in favor of DoubleTake are even larger.
Figure~\ref{fig:teach_compare} shows a qualitative comparison between the two approaches.
Table~\ref{tab:humanml_DoubleTake} presents ablations for the DoubleTake hyperparameters over the HumanML3D dataset. We show that our method using DoubleTake, soft masking, and one-second handshake size achieves the best results. Figure~\ref{fig:double_take_fail} shows qualitatively how the second take refines the first take.

\begin{table}
\centering
\resizebox{\columnwidth}{!}{

\begin{tabular}{lcccc||cc}
  \toprule
& \multicolumn{4}{c}{Motion} & \multicolumn{2}{c}{Transition} \\

\hline
&  R-precision$\uparrow$ & FID$\downarrow$ & Div.$\rightarrow$ & M.-Dist$\downarrow$ & FID$\downarrow$ & Div.$\rightarrow$\\

\hline

Ground Truth  & ${0.80}$ & $1.6 \cdot 10^{-3}$ & ${ 9.62}$ & ${2.96}$ & ${0.05}$ & ${9.57}$\\

\hline

\hline

DoubleTake (ours)  & $\underline{0.59}$ & $\mathbf{0.60}$ & $9.50$ & $5.61$ &  $\mathbf{1.48}$ & $\mathbf{8.90}$\\

\hline

First take only & $\underline{0.59}$ & ${1.00}$ & $9.46$ & $5.63$ &  ${2.15}$ & ${8.73}$\\

Second take only  & $\underline{0.59}$ & $1.09$ & $9.34$ & $\mathbf{5.57}$ &  $3.22$ & $8.35$\\

\hline

DoubleTake ($b=0$)  & $\underline{0.59}$ & ${1.00}$ & $9.51$ & $5.61$ &  $2.21$ & ${8.66}$\\

DoubleTake ($b=20$)  & $\underline{0.59}$ & $\underline{0.84}$ & $9.74$ & $\underline{5.60}$ &  $1.56$ & $8.73$\\

\hline

DoubleTake ($h=30$)  & $\mathbf{0.60}$ & ${1.03}$ & ${9.53}$ & $\underline{5.60}$ &  $2.22$ & ${8.64}$\\

DoubleTake ($h=40$)  & $0.58$ & $1.16$ & $\mathbf{9.61}$ & $5.67$ & $2.41$ & $8.61$\\

\hline

DoubleTake ($M_{soft}=0.0$)  & $\underline{0.59}$ & $0.85$ & ${9.75}$ & $5.70$ & ${1.72}$ & ${8.67}$\\

DoubleTake ($M_{soft}=0.2$)  & $\underline{0.59}$ & $0.90$ & $\underline{9.69}$ & $5.66$ & $\underline{1.50}$ & $\underline{8.77}$\\

\bottomrule
\end{tabular} 

}
\caption{\textbf{Quantitative results on the HumanML3D ~\shortcite{guo2022generating} test set.} All methods use the real motion length from the ground truth. `$\rightarrow$' means results are better if the metric is closer to the real distribution. We run all the evaluations 10 times. \textbf{Bold} indicates best result, $\underline{underline}$ indicates second best result. R-precision reported is top-3, Div. stands for diversity and M.-Dist for Multi-modal distance.}
\vspace{-15pt}
\label{tab:humanml_DoubleTake}
\end{table}

\subsection{Two-Person Generation}
\label{sec:two_person_results}

Due to the limited availability of data, we learn two-person motion in a few-shot manner. We use fixed MDM trained on the HumanML3D dataset and learn a slim communication block, ComMDM, as described in Section~\ref{sec:method}.

\textbf{Data.} 
We train and evaluate ComMDM with the 3DPW dataset \cite{von2018recovering}, which contains $27$ two-person motion sequences annotated with SMPL joints. We omit the test set since it is noisy and does not include any meaningful human interaction. Then we are left with only $10$ training examples and $4$ validation examples. Yet, the root position is often drifting which was partially fixable by reducing the drift of the camera from the drift of the root. We further augment the data by randomly mirroring and cropping each sequence. Then, we process the data to the HumanML3D joint representation, for compatibility with the original MDM input format.
We train ComMDM for two different generation tasks, both with batch size $64$ on a single NVIDIA GeForce RTX 2080 Ti GPU.

\begin{figure}[t!]
\centering
\includegraphics[width=0.9\columnwidth]{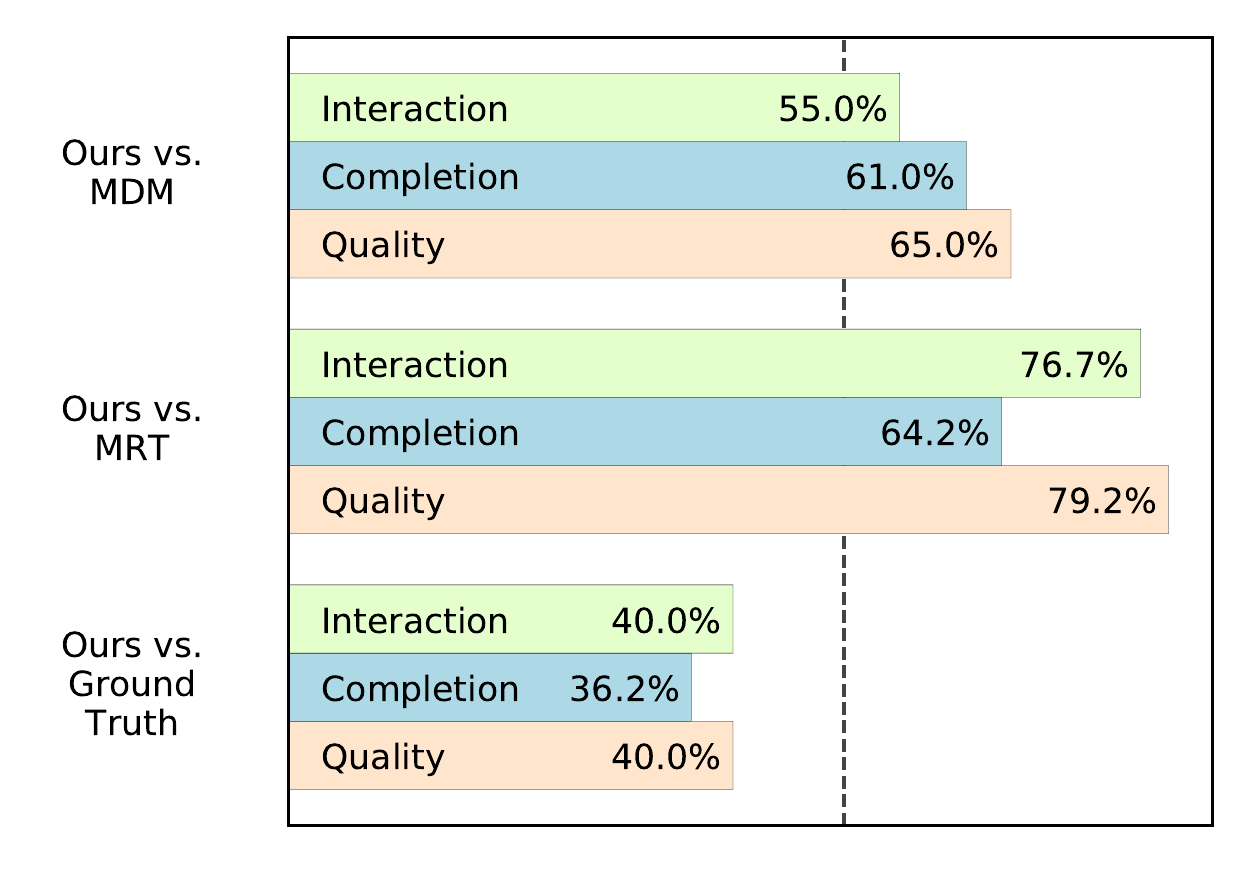}
\vspace{-15pt}
\caption{
\textbf{3DPW two-person prefix completion user study.} We asked users to compare our ComMDM to the original MDM, MRT model, and ground truth in a side-by-side view. The dashed line marks $50\%$. ComMDM outperforms both MRT and MDM in all three aspects of generation. 
}
\vspace{-10pt}
\label{fig:two_person_user_study}
\end{figure}

\textbf{Prefix completion.} We follow MRT~\cite{wang2021multi} and learn to complete 3 seconds of motion given a 1-second prefix. 
Table~\ref{tab:prefix_ablation} presents the root and joints mean $L2$ error - considering the ablation study presented, we placed the communication layer in the 8th and last layer of the transformer.
We train ComMDM for $240K$ steps.
We retrain MRT with our processed data and observe that our data process alone improved the results originally reported by the authors. 
Although MRT achieves lower error compared to our ComMDM, it generates static and unrealistic motions as presented in Figure~\ref{fig:two_person_comp}. Hence we further conducted a user study comparing ComMDM to MDM, MRT, and ground truth data, according to the aspects of \emph{interaction level}, \emph{completion of the prefix}, and \emph{overall quality} of the generated motion. $30$ unique users were participating in the user study. Each model was compared to ComMDM through $10$ randomly sampled prefixes and each such comparison was repeated by $10$ unique users. The results (Figure~\ref{fig:two_person_user_study}) show that the motions generated by ComMDM were clearly preferred over MRT and MDM.
Figure~\ref{fig:user_study_screenshot} shows an example screenshot from this user study.

\textbf{Text-to-Motion.} 
We argue that prefix completion is a motion task that becomes irrelevant. It is an explicit control signal that is both limiting the motion and giving a too-large hint for the generation.
Additionally, reporting joint error promotes dull low-frequency motion and discourages learning the distribution of motion given a condition.
Hence, we make a first step toward text control for two-person motion generation.
Since no multi-person dataset is annotated with text, we contribute 5 textual annotations for the $14$ training and validation set motions, and train ComMDM on both for $100K$ steps. Figures~\ref{fig:teaser} and \ref{fig:two_person_text} present diverse motion generation given unseen text prompts. We note that due to the small number of samples, generalization is fairly limited to interactions from the same type seen during training.

\begin{figure}[t!]
\centering
\includegraphics[width=\columnwidth]{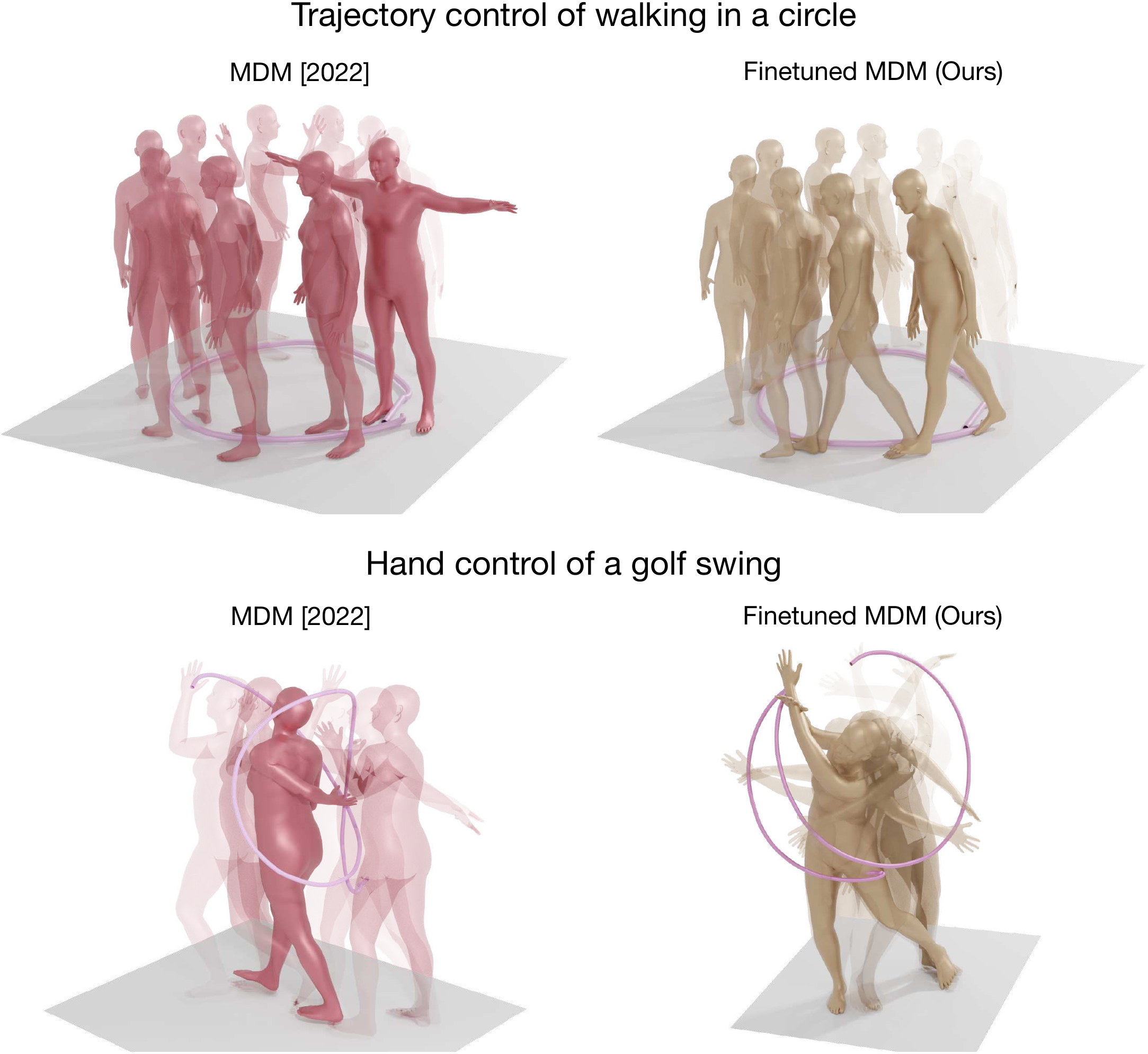}
\caption{
\textbf{Fine-tuned Motion Control (unconditioned on text)}.
We can see that MDM~\cite{tevet2023human} generates motions that completely ignore the input features: In trajectory control - MDM generates massive foot sliding, and in the hand control, the hand unrealistically bends behind the back.
Our finetuned models generate natural motions that semantically and physically match the input features: In trajectory control - we generate a walking motion that fits the trajectory and in hand control, the model recognizes the swinging motion and generates a golf swing.}
\label{fig:motion_control}
\end{figure}

\begin{table}[t!]
\centering
\resizebox{\columnwidth}{!}{

\begin{tabular}{lcccccc}
  \toprule
& \multicolumn{3}{c}{Root Error [m]} & \multicolumn{3}{c}{Joints Error [m]} \\

\hline

& $1s$ & $2s$ & $3s$ & $1s$ & $2s$ & $3s$ \\

\hline

MRT~\shortcite{wang2021multi} & $\mathbf{0.13}$ & $\mathbf{0.21}$ & $\mathbf{0.25}$ & $\mathbf{0.092}$ & $\mathbf{0.128}$& $\mathbf{0.146}$\\

\hline

ComMDM (ours)  & $\underline{0.19}$ & $\underline{0.26}$ & $\underline{0.30}$ & $\underline{0.147}$ & $\underline{0.167}$& $\underline{0.176}$\\

MDM (no Com) & $0.21$ & $0.38$ & $0.54$ & $0.162$ & $0.203$& $0.227$\\

Com only  & $0.25$ & $0.37$ & $0.47$ & $0.154$ & $0.172$& $0.181$\\

\hline

ComMDM - 2layers & $0.21$ & $0.27$ & $0.31$ & $0.162$ & $0.177$& $0.185$\\

ComMDM - 4layers & $0.22$ & $0.28$ & $0.32$ & $0.167$ & $0.182$& $0.191$\\

\hline

ComMDM @ layer6 & $0.21$ & $0.29$ & $0.33$ & $0.151$ & $0.168$& $\underline{0.176}$\\

ComMDM @ layer4 & $0.23$ & $0.32$ & $0.36$ & $0.151$ & $0.169$& $0.178$\\

ComMDM @ layer2 & $0.31$ & $0.41$ & $0.45$ & $0.156$ & $0.173$& $0.181$\\

ComMDM @ layer0 & $0.34$ & $0.44$ & $0.47$ & $0.157$ & $0.173$& $0.180$\\

\bottomrule
\end{tabular} 
}
\caption{\textbf{3DPW prefix completion L2 error.} Given a 1-second long prefix, all models predict a 3-second long motion completion. We report the root error and the joint's mean error relative to the root for the first 1, 2, and 3 seconds. \textbf{Bold} indicates best result, $\underline{\text{underline}}$ indicates second best. 
We introduce two ablation studies, the first is for the number of layers constructing ComMDM (ours is $1$), and the second is in which layer of MDM it is placed (ours is in the $8$th).
Observe that the communication block performs better when placed in higher layers of the transformer and constructed from fewer layers.}
\vspace{-10pt}
\label{tab:prefix_ablation}
\end{table}

\begin{table}[t!]
\centering
\vspace{-10pt}
\resizebox{\columnwidth}{!}{
\begin{tabular}{llccc}

\hline

\hline

& & R-precision$\uparrow$ & FID$\downarrow$ & Diversity$\rightarrow$\\

\hline

 & Ground Truth & $0.80$ & $1.6 \cdot 10^{-3}$ & $9.33$ \\

\hline

\multirow{2}{*}{\textbf{Trajectory}} &
MDM   & $0.63$ & $0.98$ & $9.04$ \\
 & Fine-tuned (Ours)  & $\mathbf{0.64}$ & $\mathbf{0.54}$ & $\mathbf{9.16}$ \\

\hline

\multirow{2}{*}{\textbf{Left Wrist}} &
MDM    & $0.63$ & $0.82$ & $\mathbf{9.31}$ \\
& Fine-tuned (Ours) & $\mathbf{0.64}$ & $\mathbf{0.34}$  & $9.41$ \\

\hline

\multirow{2}{*}{\shortstack[l]{\textbf{Left Wrist} \\ \textbf{+ Trajectory}}} &
MDM     & $0.65$ & $1.18$ & $8.81$  \\
& DiffusionBlending (Ours) & $\mathbf{0.67}$ & $\mathbf{0.22}$  & $\mathbf{9.33}$ \\

\hline

\multirow{2}{*}{\shortstack[l]{\textbf{Left Wrist} \\ \textbf{+ Right Foot}}} &
MDM     & $0.63$ & $0.81$  & $8.84$\\
& DiffusionBlending (Ours) & $\mathbf{0.67}$ & $\mathbf{0.18}$  & $\mathbf{9.35}$  \\

\bottomrule
\end{tabular} 
}
\caption{
\textbf{Joints control with fine-tuned models and DiffusionBlending.} We compare our joints control method with the motion inpainting method suggested by \citet{tevet2023human}. We conduct the evaluation on HumanML3D ~\shortcite{guo2022generating} test set. $'+'$ sign represents a blending of two fine-tuned models with our DiffusionBlending method.
}
\vspace{-20pt}
\label{tab:control}
\end{table}

\subsection{Fine-Tuned Motion Control}
\label{sec:Fine-Tuned Motion Control}
We compare our fine-tuned models and the DiffusionBlending sampling method with the original MDM model on various control tasks. 
For that sake, we sample text and control features according to each task from the HumanML3D test set. 
Motion is generated with the original MDM model by injecting the control features using the original inpainting method suggested by ~\cite{tevet2023human}. We then generate motions with the fine-tuned model that was trained for a specific control task, using our proposed inpainting method.
All fine-tuned models were initialized with the same original MDM instance we compare with, and trained with our finetuning method for $80K$ steps, with a batch size of $64$.

Note that we consider the trajectory to be the angle of the character on $xz$ plane and its linear velocities in that plane (we do not include the vertical position). 
In the joint control tasks, we take the relative location of the joint  with respect to the root location. 
For composite tasks such as left wrist+trajectory and left wrist+right foot, we apply our DiffusionBlending method on the two corresponding fine-tuned models with equal weights ($s=0.5$).  All motion control experiments were conducted above HumanML3D dataset, with text-conditioning and a classifier-free guidance scale of 2.5. Quantitative results are presented in Table~\ref{tab:control} and qualitative results are demonstrated in Figure~\ref{fig:motion_control}.
We can clearly see that fine-tuning MDM is crucial for the control task, and produces high-quality results.

\section{Conclusion}
In this paper, we have shown that a motion-based prior can be employed for advanced motion generation and control, using three novel composition methods.
We have leveraged the diffusion approach itself for the task, and have shown that it lends itself naturally to composition, enabling new tasks with little to no new data. 
Conceptually, we argue that the diffusion-based generative model can serve as a prior, or a proxy, to the human motion manifold, and thus the advanced techniques only need to address the integration between the parts being composed, relying on the fact that the generated motion is always projected back to the motion manifold. 

While promising, this initial approach is still in its infancy, and much can be further investigated. 
In long-sequence generation, for example, we are still limited to the quality of the initial model and the motion may suffer inconsistencies between distant intervals. In addition, long sequences emphasize the need to learn motions that can interact with rich environments.

In two-person motion generation, ComMDM does well at synchronizing motions between two priors, but only for interactions seen during training, lacking generalization. Based on the single-person synthesis case, we expect this approach as well to scale with larger datasets in the future. Nevertheless, two-person synthesis brings new challenges yet to be addressed.  For example, future methods should allow for valid contacts between people.

Lastly, we note the proposed techniques are not specifically designed for the motion domain. Hence perhaps the most promising avenue for future work is to adapt the techniques described in this paper (DiffusionBlending, DoubleTake, ComMDM) to other fields of generation, as well as to investigate additional ways to combine the vast knowledge embedded in pretrained generative models for novel tasks.

\section*{Acknowledgements}
We extend our gratitude to Prof. Michiel Van de Panne for his invaluable guidance, and insightful suggestions, which have significantly enriched the quality and rigor of this paper.
We thank Chuan Guo and Nikos Athanasiou for their technical support and useful advice.
We thank Sigal Raab, Roy Hachnochi and Rinon Gal for the fruitful discussions.
This research was supported in part by the Israel Science Foundation (grants no. 2492/20 and 3441/21), Len Blavatnik and the Blavatnik family foundation, and The Tel Aviv University Innovation Laboratories (TILabs). This work was supported by the Yandex Initiative in Machine Learning.

\bibliographystyle{ACM-Reference-Format}
\bibliography{MDMaP}


\begin{thebibliography}{47}


\ifx \showCODEN    \undefined \def \showCODEN     #1{\unskip}     \fi
\ifx \showDOI      \undefined \def \showDOI       #1{#1}\fi
\ifx \showISBNx    \undefined \def \showISBNx     #1{\unskip}     \fi
\ifx \showISBNxiii \undefined \def \showISBNxiii  #1{\unskip}     \fi
\ifx \showISSN     \undefined \def \showISSN      #1{\unskip}     \fi
\ifx \showLCCN     \undefined \def \showLCCN      #1{\unskip}     \fi
\ifx \shownote     \undefined \def \shownote      #1{#1}          \fi
\ifx \showarticletitle \undefined \def \showarticletitle #1{#1}   \fi
\ifx \showURL      \undefined \def \showURL       {\relax}        \fi
\providecommand\bibfield[2]{#2}
\providecommand\bibinfo[2]{#2}
\providecommand\natexlab[1]{#1}
\providecommand\showeprint[2][]{arXiv:#2}

\bibitem[CMU({[n.\,d.]})]%
        {CMU-Mocap}
 \bibinfo{year}{[n.\,d.]}\natexlab{}.
\newblock \bibinfo{title}{CMU Graphics Lab Motion Capture Database}.
\newblock \bibinfo{howpublished}{{\url{http://mocap.cs.cmu.edu/}}}.
\newblock


\bibitem[{Adobe Systems Inc.}(2021)]%
        {mixamo}
\bibfield{author}{\bibinfo{person}{{Adobe Systems Inc.}}}
  \bibinfo{year}{2021}\natexlab{}.
\newblock \bibinfo{title}{Mixamo}.
\newblock
\newblock
\urldef\tempurl%
\url{https://www.mixamo.com}
\showURL{%
\tempurl}
\newblock
\shownote{Accessed: 2021-12-25}.


\bibitem[Athanasiou et~al\mbox{.}(2022)]%
        {TEACH:3DV:2022}
\bibfield{author}{\bibinfo{person}{Nikos Athanasiou}, \bibinfo{person}{Mathis
  Petrovich}, \bibinfo{person}{Michael~J. Black}, {and}
  \bibinfo{person}{G{\"u}l Varol}.} \bibinfo{year}{2022}\natexlab{}.
\newblock \showarticletitle{{TEACH}: {T}emporal {A}ction {C}ompositions for
  {3D} {H}umans}. In \bibinfo{booktitle}{\emph{{International Conference on 3D
  Vision (3DV)}}}.
\newblock


\bibitem[Choi et~al\mbox{.}(2021)]%
        {9711284}
\bibfield{author}{\bibinfo{person}{Jooyoung Choi}, \bibinfo{person}{Sungwon
  Kim}, \bibinfo{person}{Yonghyun Jeong}, \bibinfo{person}{Youngjune Gwon},
  {and} \bibinfo{person}{Sungroh Yoon}.} \bibinfo{year}{2021}\natexlab{}.
\newblock \showarticletitle{ILVR: Conditioning Method for Denoising Diffusion
  Probabilistic Models}. In \bibinfo{booktitle}{\emph{2021 IEEE/CVF
  International Conference on Computer Vision (ICCV)}}.
  \bibinfo{pages}{14347--14356}.
\newblock
\urldef\tempurl%
\url{https://doi.org/10.1109/ICCV48922.2021.01410}
\showDOI{\tempurl}


\bibitem[Dabral et~al\mbox{.}(2023)]%
        {dabral2023mofusion}
\bibfield{author}{\bibinfo{person}{Rishabh Dabral},
  \bibinfo{person}{Muhammad~Hamza Mughal}, \bibinfo{person}{Vladislav
  Golyanik}, {and} \bibinfo{person}{Christian Theobalt}.}
  \bibinfo{year}{2023}\natexlab{}.
\newblock \showarticletitle{Mofusion: A framework for denoising-diffusion-based
  motion synthesis}. In \bibinfo{booktitle}{\emph{Proceedings of the IEEE/CVF
  Conference on Computer Vision and Pattern Recognition}}.
  \bibinfo{pages}{9760--9770}.
\newblock


\bibitem[Devlin et~al\mbox{.}(2019)]%
        {devlin-etal-2019-bert}
\bibfield{author}{\bibinfo{person}{Jacob Devlin}, \bibinfo{person}{Ming-Wei
  Chang}, \bibinfo{person}{Kenton Lee}, {and} \bibinfo{person}{Kristina
  Toutanova}.} \bibinfo{year}{2019}\natexlab{}.
\newblock \showarticletitle{{BERT}: Pre-training of Deep Bidirectional
  Transformers for Language Understanding}. In
  \bibinfo{booktitle}{\emph{Proceedings of the 2019 Conference of the North
  {A}merican Chapter of the Association for Computational Linguistics: Human
  Language Technologies, Volume 1 (Long and Short Papers)}}.
  \bibinfo{publisher}{Association for Computational Linguistics},
  \bibinfo{address}{Minneapolis, Minnesota}, \bibinfo{pages}{4171--4186}.
\newblock
\urldef\tempurl%
\url{https://doi.org/10.18653/v1/N19-1423}
\showDOI{\tempurl}


\bibitem[Guo et~al\mbox{.}(2022)]%
        {guo2022generating}
\bibfield{author}{\bibinfo{person}{Chuan Guo}, \bibinfo{person}{Shihao Zou},
  \bibinfo{person}{Xinxin Zuo}, \bibinfo{person}{Sen Wang},
  \bibinfo{person}{Wei Ji}, \bibinfo{person}{Xingyu Li}, {and}
  \bibinfo{person}{Li Cheng}.} \bibinfo{year}{2022}\natexlab{}.
\newblock \showarticletitle{Generating Diverse and Natural 3D Human Motions
  From Text}. In \bibinfo{booktitle}{\emph{Proceedings of the IEEE/CVF
  Conference on Computer Vision and Pattern Recognition}}.
  \bibinfo{pages}{5152--5161}.
\newblock


\bibitem[Ho et~al\mbox{.}(2020)]%
        {ho2020denoising}
\bibfield{author}{\bibinfo{person}{Jonathan Ho}, \bibinfo{person}{Ajay Jain},
  {and} \bibinfo{person}{Pieter Abbeel}.} \bibinfo{year}{2020}\natexlab{}.
\newblock \showarticletitle{Denoising diffusion probabilistic models}.
\newblock \bibinfo{journal}{\emph{Advances in Neural Information Processing
  Systems}}  \bibinfo{volume}{33} (\bibinfo{year}{2020}),
  \bibinfo{pages}{6840--6851}.
\newblock


\bibitem[Ho and Salimans(2022)]%
        {ho2022classifier}
\bibfield{author}{\bibinfo{person}{Jonathan Ho} {and} \bibinfo{person}{Tim
  Salimans}.} \bibinfo{year}{2022}\natexlab{}.
\newblock \showarticletitle{Classifier-free diffusion guidance}.
\newblock \bibinfo{journal}{\emph{arXiv preprint arXiv:2207.12598}}
  (\bibinfo{year}{2022}).
\newblock


\bibitem[Hong et~al\mbox{.}(2022)]%
        {hong2022avatarclip}
\bibfield{author}{\bibinfo{person}{Fangzhou Hong}, \bibinfo{person}{Mingyuan
  Zhang}, \bibinfo{person}{Liang Pan}, \bibinfo{person}{Zhongang Cai},
  \bibinfo{person}{Lei Yang}, {and} \bibinfo{person}{Ziwei Liu}.}
  \bibinfo{year}{2022}\natexlab{}.
\newblock \showarticletitle{AvatarCLIP: Zero-Shot Text-Driven Generation and
  Animation of 3D Avatars}.
\newblock \bibinfo{journal}{\emph{ACM Transactions on Graphics (TOG)}}
  \bibinfo{volume}{41}, \bibinfo{number}{4}, Article \bibinfo{articleno}{161}
  (\bibinfo{year}{2022}), \bibinfo{numpages}{19}~pages.
\newblock
\urldef\tempurl%
\url{https://doi.org/10.1145/3528223.3530094}
\showDOI{\tempurl}


\bibitem[Joo et~al\mbox{.}(2015)]%
        {Joo_2015_ICCV}
\bibfield{author}{\bibinfo{person}{Hanbyul Joo}, \bibinfo{person}{Hao Liu},
  \bibinfo{person}{Lei Tan}, \bibinfo{person}{Lin Gui}, \bibinfo{person}{Bart
  Nabbe}, \bibinfo{person}{Iain Matthews}, \bibinfo{person}{Takeo Kanade},
  \bibinfo{person}{Shohei Nobuhara}, {and} \bibinfo{person}{Yaser Sheikh}.}
  \bibinfo{year}{2015}\natexlab{}.
\newblock \showarticletitle{Panoptic Studio: A Massively Multiview System for
  Social Motion Capture}. In \bibinfo{booktitle}{\emph{The IEEE International
  Conference on Computer Vision (ICCV)}}.
\newblock


\bibitem[Karras et~al\mbox{.}(2019)]%
        {karras2019style}
\bibfield{author}{\bibinfo{person}{Tero Karras}, \bibinfo{person}{Samuli
  Laine}, {and} \bibinfo{person}{Timo Aila}.} \bibinfo{year}{2019}\natexlab{}.
\newblock \showarticletitle{A style-based generator architecture for generative
  adversarial networks}. In \bibinfo{booktitle}{\emph{Proceedings of the
  IEEE/CVF conference on computer vision and pattern recognition}}.
  \bibinfo{pages}{4401--4410}.
\newblock


\bibitem[Kim et~al\mbox{.}(2022)]%
        {kim2022flame}
\bibfield{author}{\bibinfo{person}{Jihoon Kim}, \bibinfo{person}{Jiseob Kim},
  {and} \bibinfo{person}{Sungjoon Choi}.} \bibinfo{year}{2022}\natexlab{}.
\newblock \showarticletitle{FLAME: Free-form Language-based Motion Synthesis \&
  Editing}.
\newblock \bibinfo{journal}{\emph{arXiv preprint arXiv:2209.00349}}
  (\bibinfo{year}{2022}).
\newblock


\bibitem[Kovar et~al\mbox{.}(2008)]%
        {kovar2008motion}
\bibfield{author}{\bibinfo{person}{Lucas Kovar}, \bibinfo{person}{Michael
  Gleicher}, {and} \bibinfo{person}{Fr{\'e}d{\'e}ric Pighin}.}
  \bibinfo{year}{2008}\natexlab{}.
\newblock \showarticletitle{Motion graphs}.
\newblock In \bibinfo{booktitle}{\emph{ACM SIGGRAPH 2008 classes}}.
  \bibinfo{pages}{1--10}.
\newblock


\bibitem[Loper et~al\mbox{.}(2015)]%
        {loper2015smpl}
\bibfield{author}{\bibinfo{person}{Matthew Loper}, \bibinfo{person}{Naureen
  Mahmood}, \bibinfo{person}{Javier Romero}, \bibinfo{person}{Gerard
  Pons-Moll}, {and} \bibinfo{person}{Michael~J Black}.}
  \bibinfo{year}{2015}\natexlab{}.
\newblock \showarticletitle{SMPL: A skinned multi-person linear model}.
\newblock \bibinfo{journal}{\emph{ACM transactions on graphics (TOG)}}
  \bibinfo{volume}{34}, \bibinfo{number}{6} (\bibinfo{year}{2015}),
  \bibinfo{pages}{1--16}.
\newblock


\bibitem[Lugmayr et~al\mbox{.}(2022)]%
        {lugmayr2022repaint}
\bibfield{author}{\bibinfo{person}{Andreas Lugmayr}, \bibinfo{person}{Martin
  Danelljan}, \bibinfo{person}{Andres Romero}, \bibinfo{person}{Fisher Yu},
  \bibinfo{person}{Radu Timofte}, {and} \bibinfo{person}{Luc Van~Gool}.}
  \bibinfo{year}{2022}\natexlab{}.
\newblock \showarticletitle{Repaint: Inpainting using denoising diffusion
  probabilistic models}. In \bibinfo{booktitle}{\emph{Proceedings of the
  IEEE/CVF Conference on Computer Vision and Pattern Recognition}}.
  \bibinfo{pages}{11461--11471}.
\newblock


\bibitem[Mahmood et~al\mbox{.}(2019)]%
        {AMASS:ICCV:2019}
\bibfield{author}{\bibinfo{person}{Naureen Mahmood}, \bibinfo{person}{Nima
  Ghorbani}, \bibinfo{person}{Nikolaus~F. Troje}, \bibinfo{person}{Gerard
  Pons-Moll}, {and} \bibinfo{person}{Michael~J. Black}.}
  \bibinfo{year}{2019}\natexlab{}.
\newblock \showarticletitle{{AMASS}: Archive of Motion Capture as Surface
  Shapes}. In \bibinfo{booktitle}{\emph{International Conference on Computer
  Vision}}. \bibinfo{pages}{5442--5451}.
\newblock


\bibitem[Mao et~al\mbox{.}(2022)]%
        {mao2022weakly}
\bibfield{author}{\bibinfo{person}{Wei Mao}, \bibinfo{person}{Miaomiao Liu},
  {and} \bibinfo{person}{Mathieu Salzmann}.} \bibinfo{year}{2022}\natexlab{}.
\newblock \showarticletitle{Weakly-supervised Action Transition Learning for
  Stochastic Human Motion Prediction}. In \bibinfo{booktitle}{\emph{Proceedings
  of the IEEE/CVF Conference on Computer Vision and Pattern Recognition}}.
  \bibinfo{pages}{8151--8160}.
\newblock


\bibitem[Martinez et~al\mbox{.}(2017)]%
        {martinez2017human}
\bibfield{author}{\bibinfo{person}{Julieta Martinez},
  \bibinfo{person}{Michael~J Black}, {and} \bibinfo{person}{Javier Romero}.}
  \bibinfo{year}{2017}\natexlab{}.
\newblock \showarticletitle{On human motion prediction using recurrent neural
  networks}. In \bibinfo{booktitle}{\emph{Proceedings of the IEEE conference on
  computer vision and pattern recognition}}. \bibinfo{pages}{2891--2900}.
\newblock


\bibitem[Mehta et~al\mbox{.}(2018)]%
        {mehta2018single}
\bibfield{author}{\bibinfo{person}{Dushyant Mehta}, \bibinfo{person}{Oleksandr
  Sotnychenko}, \bibinfo{person}{Franziska Mueller}, \bibinfo{person}{Weipeng
  Xu}, \bibinfo{person}{Srinath Sridhar}, \bibinfo{person}{Gerard Pons-Moll},
  {and} \bibinfo{person}{Christian Theobalt}.} \bibinfo{year}{2018}\natexlab{}.
\newblock \showarticletitle{Single-shot multi-person 3d pose estimation from
  monocular rgb}. In \bibinfo{booktitle}{\emph{2018 International Conference on
  3D Vision (3DV)}}. IEEE, \bibinfo{pages}{120--130}.
\newblock


\bibitem[Meng et~al\mbox{.}(2022)]%
        {meng2022sdedit}
\bibfield{author}{\bibinfo{person}{Chenlin Meng}, \bibinfo{person}{Yutong He},
  \bibinfo{person}{Yang Song}, \bibinfo{person}{Jiaming Song},
  \bibinfo{person}{Jiajun Wu}, \bibinfo{person}{Jun-Yan Zhu}, {and}
  \bibinfo{person}{Stefano Ermon}.} \bibinfo{year}{2022}\natexlab{}.
\newblock \showarticletitle{{SDE}dit: Guided Image Synthesis and Editing with
  Stochastic Differential Equations}. In
  \bibinfo{booktitle}{\emph{International Conference on Learning
  Representations}}.
\newblock


\bibitem[Pavlakos et~al\mbox{.}(2019)]%
        {SMPL-X:2019}
\bibfield{author}{\bibinfo{person}{Georgios Pavlakos},
  \bibinfo{person}{Vasileios Choutas}, \bibinfo{person}{Nima Ghorbani},
  \bibinfo{person}{Timo Bolkart}, \bibinfo{person}{Ahmed A.~A. Osman},
  \bibinfo{person}{Dimitrios Tzionas}, {and} \bibinfo{person}{Michael~J.
  Black}.} \bibinfo{year}{2019}\natexlab{}.
\newblock \showarticletitle{Expressive Body Capture: {3D} Hands, Face, and Body
  from a Single Image}. In \bibinfo{booktitle}{\emph{Proceedings IEEE Conf. on
  Computer Vision and Pattern Recognition (CVPR)}}.
  \bibinfo{pages}{10975--10985}.
\newblock


\bibitem[Petrovich et~al\mbox{.}(2022)]%
        {petrovich22temos}
\bibfield{author}{\bibinfo{person}{Mathis Petrovich},
  \bibinfo{person}{Michael~J. Black}, {and} \bibinfo{person}{G{\"u}l Varol}.}
  \bibinfo{year}{2022}\natexlab{}.
\newblock \showarticletitle{{TEMOS}: Generating diverse human motions from
  textual descriptions}. In \bibinfo{booktitle}{\emph{European Conference on
  Computer Vision ({ECCV})}}.
\newblock


\bibitem[Punnakkal et~al\mbox{.}(2021)]%
        {BABEL:CVPR:2021}
\bibfield{author}{\bibinfo{person}{Abhinanda~R. Punnakkal},
  \bibinfo{person}{Arjun Chandrasekaran}, \bibinfo{person}{Nikos Athanasiou},
  \bibinfo{person}{Alejandra Quiros-Ramirez}, {and} \bibinfo{person}{Michael~J.
  Black}.} \bibinfo{year}{2021}\natexlab{}.
\newblock \showarticletitle{{BABEL}: Bodies, Action and Behavior with English
  Labels}. In \bibinfo{booktitle}{\emph{Proceedings IEEE/CVF Conf.~on Computer
  Vision and Pattern Recognition (CVPR)}}. \bibinfo{pages}{722--731}.
\newblock


\bibitem[Raab et~al\mbox{.}(2022)]%
        {raab2022modi}
\bibfield{author}{\bibinfo{person}{Sigal Raab}, \bibinfo{person}{Inbal
  Leibovitch}, \bibinfo{person}{Peizhuo Li}, \bibinfo{person}{Kfir Aberman},
  \bibinfo{person}{Olga Sorkine-Hornung}, {and} \bibinfo{person}{Daniel
  Cohen-Or}.} \bibinfo{year}{2022}\natexlab{}.
\newblock \showarticletitle{MoDi: Unconditional Motion Synthesis from Diverse
  Data}.
\newblock \bibinfo{journal}{\emph{arXiv preprint arXiv:2206.08010}}
  (\bibinfo{year}{2022}).
\newblock


\bibitem[Raab et~al\mbox{.}(2023)]%
        {raab2023single}
\bibfield{author}{\bibinfo{person}{Sigal Raab}, \bibinfo{person}{Inbal
  Leibovitch}, \bibinfo{person}{Guy Tevet}, \bibinfo{person}{Moab Arar},
  \bibinfo{person}{Amit~H Bermano}, {and} \bibinfo{person}{Daniel Cohen-Or}.}
  \bibinfo{year}{2023}\natexlab{}.
\newblock \showarticletitle{Single Motion Diffusion}.
\newblock \bibinfo{journal}{\emph{arXiv preprint arXiv:2302.05905}}
  (\bibinfo{year}{2023}).
\newblock


\bibitem[Radford et~al\mbox{.}(2021)]%
        {radford2021learning}
\bibfield{author}{\bibinfo{person}{Alec Radford}, \bibinfo{person}{Jong~Wook
  Kim}, \bibinfo{person}{Chris Hallacy}, \bibinfo{person}{Aditya Ramesh},
  \bibinfo{person}{Gabriel Goh}, \bibinfo{person}{Sandhini Agarwal},
  \bibinfo{person}{Girish Sastry}, \bibinfo{person}{Amanda Askell},
  \bibinfo{person}{Pamela Mishkin}, \bibinfo{person}{Jack Clark},
  {et~al\mbox{.}}} \bibinfo{year}{2021}\natexlab{}.
\newblock \showarticletitle{Learning transferable visual models from natural
  language supervision}. In \bibinfo{booktitle}{\emph{International Conference
  on Machine Learning}}. PMLR, \bibinfo{pages}{8748--8763}.
\newblock


\bibitem[Rombach et~al\mbox{.}(2022a)]%
        {Rombach_2022_CVPR}
\bibfield{author}{\bibinfo{person}{Robin Rombach}, \bibinfo{person}{Andreas
  Blattmann}, \bibinfo{person}{Dominik Lorenz}, \bibinfo{person}{Patrick
  Esser}, {and} \bibinfo{person}{Bj\"orn Ommer}.}
  \bibinfo{year}{2022}\natexlab{a}.
\newblock \showarticletitle{High-Resolution Image Synthesis With Latent
  Diffusion Models}. In \bibinfo{booktitle}{\emph{Proceedings of the IEEE/CVF
  Conference on Computer Vision and Pattern Recognition (CVPR)}}.
  \bibinfo{pages}{10684--10695}.
\newblock


\bibitem[Rombach et~al\mbox{.}(2022b)]%
        {rombach2022high}
\bibfield{author}{\bibinfo{person}{Robin Rombach}, \bibinfo{person}{Andreas
  Blattmann}, \bibinfo{person}{Dominik Lorenz}, \bibinfo{person}{Patrick
  Esser}, {and} \bibinfo{person}{Bj{\"o}rn Ommer}.}
  \bibinfo{year}{2022}\natexlab{b}.
\newblock \showarticletitle{High-resolution image synthesis with latent
  diffusion models}. In \bibinfo{booktitle}{\emph{Proceedings of the IEEE/CVF
  Conference on Computer Vision and Pattern Recognition}}.
  \bibinfo{pages}{10684--10695}.
\newblock


\bibitem[Saharia et~al\mbox{.}(2022)]%
        {saharia2022palette}
\bibfield{author}{\bibinfo{person}{Chitwan Saharia}, \bibinfo{person}{William
  Chan}, \bibinfo{person}{Huiwen Chang}, \bibinfo{person}{Chris Lee},
  \bibinfo{person}{Jonathan Ho}, \bibinfo{person}{Tim Salimans},
  \bibinfo{person}{David Fleet}, {and} \bibinfo{person}{Mohammad Norouzi}.}
  \bibinfo{year}{2022}\natexlab{}.
\newblock \showarticletitle{Palette: Image-to-image diffusion models}. In
  \bibinfo{booktitle}{\emph{ACM SIGGRAPH 2022 Conference Proceedings}}.
  \bibinfo{pages}{1--10}.
\newblock


\bibitem[Sohl-Dickstein et~al\mbox{.}(2015)]%
        {sohl2015deep}
\bibfield{author}{\bibinfo{person}{Jascha Sohl-Dickstein},
  \bibinfo{person}{Eric Weiss}, \bibinfo{person}{Niru Maheswaranathan}, {and}
  \bibinfo{person}{Surya Ganguli}.} \bibinfo{year}{2015}\natexlab{}.
\newblock \showarticletitle{Deep unsupervised learning using nonequilibrium
  thermodynamics}. In \bibinfo{booktitle}{\emph{International Conference on
  Machine Learning}}. PMLR, \bibinfo{pages}{2256--2265}.
\newblock


\bibitem[Song et~al\mbox{.}(2020)]%
        {song2020score}
\bibfield{author}{\bibinfo{person}{Yang Song}, \bibinfo{person}{Jascha
  Sohl-Dickstein}, \bibinfo{person}{Diederik~P Kingma},
  \bibinfo{person}{Abhishek Kumar}, \bibinfo{person}{Stefano Ermon}, {and}
  \bibinfo{person}{Ben Poole}.} \bibinfo{year}{2020}\natexlab{}.
\newblock \showarticletitle{Score-based generative modeling through stochastic
  differential equations}.
\newblock \bibinfo{journal}{\emph{arXiv preprint arXiv:2011.13456}}
  (\bibinfo{year}{2020}).
\newblock


\bibitem[Song et~al\mbox{.}(2022)]%
        {song2022actformer}
\bibfield{author}{\bibinfo{person}{Ziyang Song}, \bibinfo{person}{Dongliang
  Wang}, \bibinfo{person}{Nan Jiang}, \bibinfo{person}{Zhicheng Fang},
  \bibinfo{person}{Chenjing Ding}, \bibinfo{person}{Weihao Gan}, {and}
  \bibinfo{person}{Wei Wu}.} \bibinfo{year}{2022}\natexlab{}.
\newblock \showarticletitle{ActFormer: A GAN Transformer Framework towards
  General Action-Conditioned 3D Human Motion Generation}.
\newblock \bibinfo{journal}{\emph{arXiv preprint arXiv:2203.07706}}
  (\bibinfo{year}{2022}).
\newblock


\bibitem[Tevet et~al\mbox{.}(2022)]%
        {tevet2022motionclip}
\bibfield{author}{\bibinfo{person}{Guy Tevet}, \bibinfo{person}{Brian Gordon},
  \bibinfo{person}{Amir Hertz}, \bibinfo{person}{Amit~H Bermano}, {and}
  \bibinfo{person}{Daniel Cohen-Or}.} \bibinfo{year}{2022}\natexlab{}.
\newblock \showarticletitle{Motionclip: Exposing human motion generation to
  clip space}. In \bibinfo{booktitle}{\emph{Computer Vision--ECCV 2022: 17th
  European Conference, Tel Aviv, Israel, October 23--27, 2022, Proceedings,
  Part XXII}}. Springer, \bibinfo{pages}{358--374}.
\newblock


\bibitem[Tevet et~al\mbox{.}(2023)]%
        {tevet2023human}
\bibfield{author}{\bibinfo{person}{Guy Tevet}, \bibinfo{person}{Sigal Raab},
  \bibinfo{person}{Brian Gordon}, \bibinfo{person}{Yoni Shafir},
  \bibinfo{person}{Daniel Cohen-or}, {and} \bibinfo{person}{Amit~Haim
  Bermano}.} \bibinfo{year}{2023}\natexlab{}.
\newblock \showarticletitle{Human Motion Diffusion Model}. In
  \bibinfo{booktitle}{\emph{The Eleventh International Conference on Learning
  Representations}}.
\newblock
\urldef\tempurl%
\url{https://openreview.net/forum?id=SJ1kSyO2jwu}
\showURL{%
\tempurl}


\bibitem[Tiwari et~al\mbox{.}(2022)]%
        {tiwari2022pose}
\bibfield{author}{\bibinfo{person}{Garvita Tiwari}, \bibinfo{person}{Dimitrije
  Anti{\'c}}, \bibinfo{person}{Jan~Eric Lenssen}, \bibinfo{person}{Nikolaos
  Sarafianos}, \bibinfo{person}{Tony Tung}, {and} \bibinfo{person}{Gerard
  Pons-Moll}.} \bibinfo{year}{2022}\natexlab{}.
\newblock \showarticletitle{Pose-ndf: Modeling human pose manifolds with neural
  distance fields}. In \bibinfo{booktitle}{\emph{European Conference on
  Computer Vision}}. Springer, \bibinfo{pages}{572--589}.
\newblock


\bibitem[Tseng et~al\mbox{.}(2022)]%
        {tseng2022edge}
\bibfield{author}{\bibinfo{person}{Jonathan Tseng}, \bibinfo{person}{Rodrigo
  Castellon}, {and} \bibinfo{person}{C~Karen Liu}.}
  \bibinfo{year}{2022}\natexlab{}.
\newblock \showarticletitle{EDGE: Editable Dance Generation From Music}.
\newblock \bibinfo{journal}{\emph{arXiv preprint arXiv:2211.10658}}
  (\bibinfo{year}{2022}).
\newblock


\bibitem[Vendrow et~al\mbox{.}(2022)]%
        {vendrow2022somoformer}
\bibfield{author}{\bibinfo{person}{Edward Vendrow}, \bibinfo{person}{Satyajit
  Kumar}, \bibinfo{person}{Ehsan Adeli}, {and} \bibinfo{person}{Hamid
  Rezatofighi}.} \bibinfo{year}{2022}\natexlab{}.
\newblock \showarticletitle{SoMoFormer: Multi-Person Pose Forecasting with
  Transformers}.
\newblock \bibinfo{journal}{\emph{arXiv preprint arXiv:2208.14023}}
  (\bibinfo{year}{2022}).
\newblock


\bibitem[Von~Marcard et~al\mbox{.}(2018)]%
        {von2018recovering}
\bibfield{author}{\bibinfo{person}{Timo Von~Marcard}, \bibinfo{person}{Roberto
  Henschel}, \bibinfo{person}{Michael~J Black}, \bibinfo{person}{Bodo
  Rosenhahn}, {and} \bibinfo{person}{Gerard Pons-Moll}.}
  \bibinfo{year}{2018}\natexlab{}.
\newblock \showarticletitle{Recovering accurate 3d human pose in the wild using
  imus and a moving camera}. In \bibinfo{booktitle}{\emph{Proceedings of the
  European Conference on Computer Vision (ECCV)}}. \bibinfo{pages}{601--617}.
\newblock


\bibitem[Wang et~al\mbox{.}(2021)]%
        {wang2021multi}
\bibfield{author}{\bibinfo{person}{Jiashun Wang}, \bibinfo{person}{Huazhe Xu},
  \bibinfo{person}{Medhini Narasimhan}, {and} \bibinfo{person}{Xiaolong Wang}.}
  \bibinfo{year}{2021}\natexlab{}.
\newblock \showarticletitle{Multi-Person 3D Motion Prediction with Multi-Range
  Transformers}.
\newblock \bibinfo{journal}{\emph{Advances in Neural Information Processing
  Systems}}  \bibinfo{volume}{34} (\bibinfo{year}{2021}).
\newblock


\bibitem[Wang et~al\mbox{.}(2022)]%
        {wang2022neural}
\bibfield{author}{\bibinfo{person}{Weiqiang Wang}, \bibinfo{person}{Xuefei
  Zhe}, \bibinfo{person}{Huan Chen}, \bibinfo{person}{Di Kang},
  \bibinfo{person}{Tingguang Li}, \bibinfo{person}{Ruizhi Chen}, {and}
  \bibinfo{person}{Linchao Bao}.} \bibinfo{year}{2022}\natexlab{}.
\newblock \showarticletitle{NEURAL MARIONETTE: A Transformer-based Multi-action
  Human Motion Synthesis System}.
\newblock \bibinfo{journal}{\emph{arXiv preprint arXiv:2209.13204}}
  (\bibinfo{year}{2022}).
\newblock


\bibitem[Wu et~al\mbox{.}(2021)]%
        {wu2021stylealign}
\bibfield{author}{\bibinfo{person}{Zongze Wu}, \bibinfo{person}{Yotam Nitzan},
  \bibinfo{person}{Eli Shechtman}, {and} \bibinfo{person}{Dani Lischinski}.}
  \bibinfo{year}{2021}\natexlab{}.
\newblock \showarticletitle{StyleAlign: Analysis and Applications of Aligned
  StyleGAN Models}.
\newblock \bibinfo{journal}{\emph{arXiv preprint arXiv:2110.11323}}
  (\bibinfo{year}{2021}).
\newblock


\bibitem[Xin et~al\mbox{.}(2022)]%
        {chen2022mld}
\bibfield{author}{\bibinfo{person}{Chen Xin}, \bibinfo{person}{Biao Jiang},
  \bibinfo{person}{Wen Liu}, \bibinfo{person}{Zilong Huang},
  \bibinfo{person}{Bin Fu}, \bibinfo{person}{Tao Chen}, \bibinfo{person}{Jingyi
  Yu}, {and} \bibinfo{person}{Gang Yu}.} \bibinfo{year}{2022}\natexlab{}.
\newblock \showarticletitle{Executing your Commands via Motion Diffusion in
  Latent Space}.
\newblock \bibinfo{journal}{\emph{arXiv}} (\bibinfo{year}{2022}).
\newblock


\bibitem[Yin et~al\mbox{.}(2018)]%
        {yin2018sampling}
\bibfield{author}{\bibinfo{person}{Kangxue Yin}, \bibinfo{person}{Hui Huang},
  \bibinfo{person}{Edmond~SL Ho}, \bibinfo{person}{Hao Wang},
  \bibinfo{person}{Taku Komura}, \bibinfo{person}{Daniel Cohen-Or}, {and}
  \bibinfo{person}{Hao Zhang}.} \bibinfo{year}{2018}\natexlab{}.
\newblock \showarticletitle{A sampling approach to generating closely
  interacting 3d pose-pairs from 2d annotations}.
\newblock \bibinfo{journal}{\emph{IEEE transactions on visualization and
  computer graphics}} \bibinfo{volume}{25}, \bibinfo{number}{6}
  (\bibinfo{year}{2018}), \bibinfo{pages}{2217--2227}.
\newblock


\bibitem[Yuan et~al\mbox{.}(2022)]%
        {yuan2022physdiff}
\bibfield{author}{\bibinfo{person}{Ye Yuan}, \bibinfo{person}{Jiaming Song},
  \bibinfo{person}{Umar Iqbal}, \bibinfo{person}{Arash Vahdat}, {and}
  \bibinfo{person}{Jan Kautz}.} \bibinfo{year}{2022}\natexlab{}.
\newblock \showarticletitle{PhysDiff: Physics-Guided Human Motion Diffusion
  Model}.
\newblock \bibinfo{journal}{\emph{arXiv preprint arXiv:2212.02500}}
  (\bibinfo{year}{2022}).
\newblock


\bibitem[Zhang et~al\mbox{.}(2022)]%
        {zhang2022motiondiffuse}
\bibfield{author}{\bibinfo{person}{Mingyuan Zhang}, \bibinfo{person}{Zhongang
  Cai}, \bibinfo{person}{Liang Pan}, \bibinfo{person}{Fangzhou Hong},
  \bibinfo{person}{Xinying Guo}, \bibinfo{person}{Lei Yang}, {and}
  \bibinfo{person}{Ziwei Liu}.} \bibinfo{year}{2022}\natexlab{}.
\newblock \showarticletitle{MotionDiffuse: Text-Driven Human Motion Generation
  with Diffusion Model}.
\newblock \bibinfo{journal}{\emph{arXiv preprint arXiv:2208.15001}}
  (\bibinfo{year}{2022}).
\newblock


\bibitem[Zhou et~al\mbox{.}(2018)]%
        {zhou2018auto}
\bibfield{author}{\bibinfo{person}{Yi Zhou}, \bibinfo{person}{Zimo Li},
  \bibinfo{person}{Shuangjiu Xiao}, \bibinfo{person}{Chong He},
  \bibinfo{person}{Zeng Huang}, {and} \bibinfo{person}{Hao Li}.}
  \bibinfo{year}{2018}\natexlab{}.
\newblock \showarticletitle{Auto-Conditioned Recurrent Networks for Extended
  Complex Human Motion Synthesis}. In \bibinfo{booktitle}{\emph{International
  Conference on Learning Representations}}.
\newblock


\end{thebibliography}

\appendix
\section{User Study}
\label{sec:user_study}

\begin{figure}[t!]
\centering
\includegraphics[width=0.9\columnwidth]{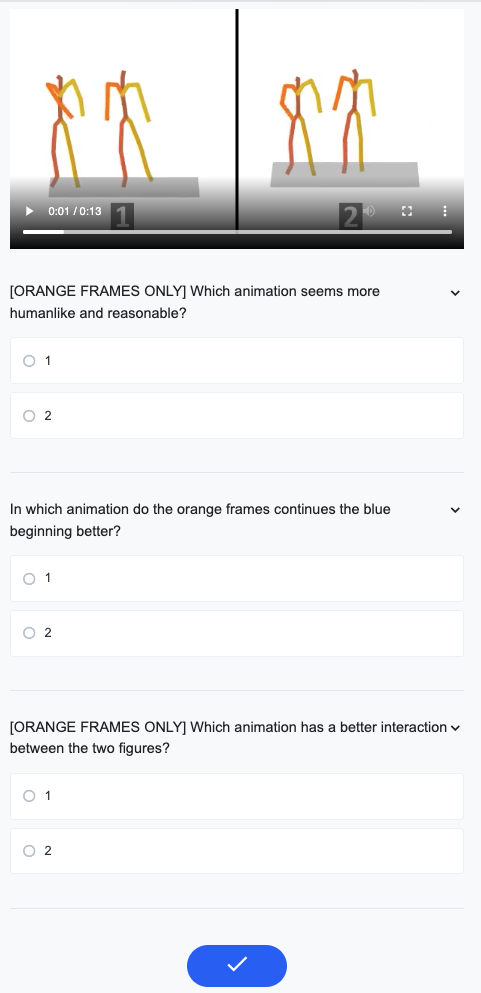}
\caption{
A sample screenshot from the two-person prefix completion user study.
}
\label{fig:user_study_screenshot}
\end{figure}

We conducted a user study of the two-person prefix completion task. Its details can be found in Section~\ref{sec:two_person_results} 
and the results are presented in Figure~\ref{fig:two_person_user_study}. Figure~\ref{fig:user_study_screenshot} presents a sample screenshot from the user study form.

\clearpage

\end{document}